\documentclass[11pt]{article}

\usepackage{latex/acl}

\usepackage{times}
\usepackage{latexsym}
\usepackage[table]{xcolor}
\usepackage{makecell}

\usepackage[T1]{fontenc}

\usepackage[utf8]{inputenc}

\usepackage{microtype}

\usepackage{inconsolata}

\usepackage{graphicx}

\usepackage{amsmath}
\usepackage{amssymb}
\usepackage{booktabs}
\usepackage{algorithm}
\usepackage{tabularx}
\usepackage{pgfplots}
\usepackage{pgfplotstable}
\usepgfplotslibrary{fillbetween}
\pgfplotsset{compat=1.18}
\usepackage{multirow}
\usepackage[l3]{csvsimple}
\usepackage{tcolorbox}

\usepackage{subcaption}

\usepackage{pgfplots}

\usepackage[font=small]{caption}

\usepackage[normalem]{ulem}
\usepackage{xcolor}

\newcommand{\condppls}[0]{Uncertainty}

\newcommand{\srcentrerank}[0]{Source Entropy Reranking}

\newcommand{\rankragrerank}[0]{RankRAG Reranking}

\newcommand{\logitgaprerank}[0]{TARG Reranking}

\newcommand{\condpplrerank}[0]{Conditional Entropy Rerank}

\newcommand{\ayas}[0]{Aya 8B}

\newcommand{\crbs}[0]{Command R7B}

\newcommand{\qwenss}[0]{Qwen 3B}

\newcommand{\mistrals}[0]{Mistral 8B}
\newcommand{\llamas}[0]{Llama-3.2-3B-Instruct}
\newcommand{\llamass}[0]{Llama 3B}

\newcommand{\llamams}[0]{Llama 8B}

\newcommand{\qwenbs}[0]{Qwen 32B}

\newcommand{\llamabs}[0]{Llama 70B}

\title{When Reranking Hurts:\\ 
Uncertainty-Based Gating for Few-Shot Reranking}

\author{
  Orian Dabod\textsuperscript{1}, 
  Amir DN Cohen\textsuperscript{2}, 
  Gabriel Stanovsky\textsuperscript{1} \\
  \\
  \textsuperscript{1}The Hebrew University of Jerusalem, 
  \textsuperscript{2}OriginAI / Ramat Gan, Israel \\
  \texttt{orian.dabod@mail.huji.ac.il}
}

\usepackage[normalem]{ulem}
\usepackage{xcolor}

\newcommand{\com}[1]{}

\newcommand{\resolved}[1]{}

\begin{document}
\maketitle

\begin{abstract}
Few-shot selection typically assumes that reranking retrieved examples always improves performance. We challenge this view by identifying that the expensive reranking step can in fact degrade performance. Instead, we propose \emph{Training-Free Gated Reranking}, which decides whether to rerank the few-shot examples based on the model's uncertainty. Extensive experiments across 8 LLMs, covering 7 NLU datasets and 18 MT domain-direction combinations, demonstrate that our approach reduces average computational costs by approximately 20\% for MT and 54\% for NLU at a calibrated operating point, while matching the performance of full reranking. These findings indicate that higher computational cost does not guarantee better performance, and that reranking is most beneficial when targeted at high-uncertainty instances.
\end{abstract}

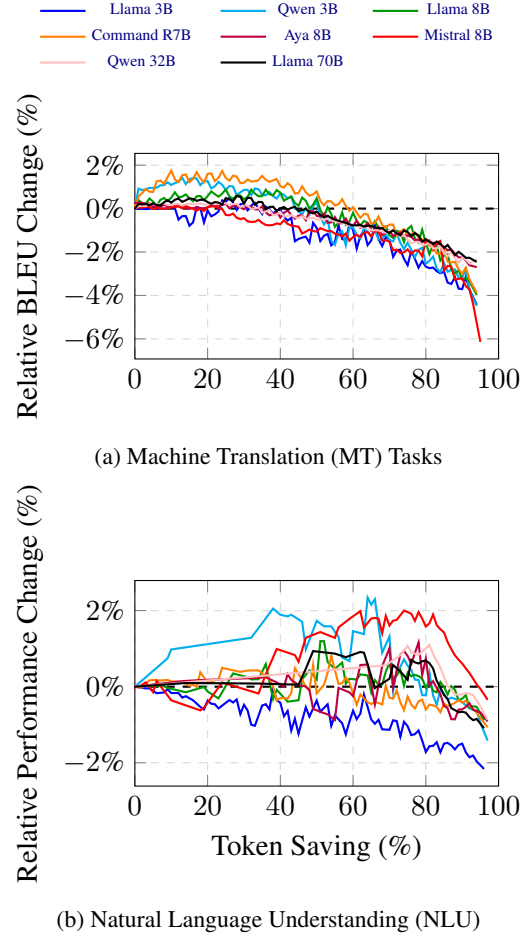
\begin{figure}[t]
    \centering
    \ref{shared_legend}
    \begin{subfigure}[b]{\linewidth}
        \centering
        \begin{tikzpicture}
            \begin{axis}[
                width=0.83\linewidth,
                height=4.3cm,
                legend to name=shared_legend, 
                legend columns=3,             
                legend style={
                    font=\tiny, 
                    draw=none,                
                    /tikz/every even column/.append style={column sep=0.3cm}
                },
                ylabel={Relative BLEU Change (\%)},
                xmin=0, xmax=100,
                yticklabel={\pgfmathprintnumber{\tick}\%},
                grid=major,
                grid style={dashed, gray!30},
                title style={font=\bfseries},
                cycle list name=color list
            ]
                \addplot[black, dashed, thick, forget plot] coordinates {(0,0) (100,0)};

                \addplot[blue, thick] table [x=savings_percent, y=mean_smooth, col sep=comma] {graphs/unsloth_Llama-3.2-3B-Instruct_graph_data.csv};
                \addlegendentry{\llamass{}}

                \addplot[cyan, thick] table [x=savings_percent, y=mean_smooth, col sep=comma] {graphs/Qwen_Qwen2.5-3B-Instruct_graph_data.csv};
                \addlegendentry{\qwenss{}}

                \addplot[green!60!black, thick] table [x=savings_percent, y=mean_smooth, col sep=comma] {graphs/unsloth_Meta-Llama-3.1-8B-Instruct_graph_data.csv};
                \addlegendentry{\llamams{}}

                \addplot[orange, thick] table [x=savings_percent, y=mean_smooth, col sep=comma] {graphs/CohereLabs_c4ai-command-r7b-12-2024_graph_data.csv};
                \addlegendentry{\crbs{}}

                \addplot[purple, thick] table [x=savings_percent, y=mean_smooth, col sep=comma] {graphs/CohereLabs_aya-expanse-8b_graph_data.csv};
                \addlegendentry{\ayas{}}

                \addplot[red, thick] table [x=savings_percent, y=mean_smooth, col sep=comma] {graphs/mistralai_Mistral-7B-Instruct-v0.3_graph_data.csv};
                \addlegendentry{\mistrals{}}

                \addplot[pink, thick] table [x=savings_percent, y=mean_smooth, col sep=comma] {graphs/Qwen_Qwen2.5-32B-Instruct_graph_data.csv};
                \addlegendentry{\qwenbs{}}

                \addplot[black, thick] table [x=savings_percent, y=mean_smooth, col sep=comma] {graphs/unsloth_Llama-3.3-70B-Instruct_graph_data.csv};
                \addlegendentry{\llamabs{}}
            \end{axis}
        \end{tikzpicture}
        \caption{Machine Translation (MT) Tasks}
        \label{fig:model_comparison_mt}
    \end{subfigure}
        
    \begin{subfigure}[b]{\linewidth}
        \centering
        \begin{tikzpicture}
            \begin{axis}[
                width=0.83\linewidth,
                height=4.3cm,
                xlabel={Token Saving (\%)}, %
                ylabel={Relative Performance Change (\%)}, 
                xmin=0, xmax=100,
                yticklabel={\pgfmathprintnumber{\tick}\%},
                grid=major,
                grid style={dashed, gray!30},
                title style={font=\bfseries},
                cycle list name=color list
            ]
                \addplot[black, dashed, thick, forget plot] coordinates {(0,0) (100,0)};

                \addplot[blue, thick] table [x=savings_percent, y=mean_smooth, col sep=comma] {graphs/nlu/unsloth_Llama-3.2-3B-Instruct_graph_data.csv};
                
                \addplot[cyan, thick] table [x=savings_percent, y=mean_smooth, col sep=comma] {graphs/nlu/Qwen_Qwen2.5-3B-Instruct_graph_data.csv};
                
                \addplot[green!60!black, thick] table [x=savings_percent, y=mean_smooth, col sep=comma] {graphs/nlu/unsloth_Meta-Llama-3.1-8B-Instruct_graph_data.csv};
                
                \addplot[orange, thick] table [x=savings_percent, y=mean_smooth, col sep=comma] {graphs/nlu/CohereLabs_c4ai-command-r7b-12-2024_graph_data.csv};
                
                \addplot[purple, thick] table [x=savings_percent, y=mean_smooth, col sep=comma] {graphs/nlu/CohereLabs_aya-expanse-8b_graph_data.csv};
                
                \addplot[red, thick] table [x=savings_percent, y=mean_smooth, col sep=comma] {graphs/nlu/mistralai_Mistral-7B-Instruct-v0.3_graph_data.csv};
                
                \addplot[pink, thick] table [x=savings_percent, y=mean_smooth, col sep=comma] {graphs/nlu/Qwen_Qwen2.5-32B-Instruct_graph_data.csv};
                
                \addplot[black, thick] table [x=savings_percent, y=mean_smooth, col sep=comma] {graphs/nlu/unsloth_Llama-3.3-70B-Instruct_graph_data.csv};
                
            \end{axis}
        \end{tikzpicture}
        \caption{Natural Language Understanding (NLU)}
        \label{fig:model_comparison_nlu}
    \end{subfigure}
    
    \caption{Relative performance impact versus computational token savings achieved via selective reranking for Machine Translation (MT, measured in BLEU) and Natural Language Understanding (NLU) tasks. The majority of models (3B--70B) exhibit performance gains, though the optimal savings window shifts from 15--50\% for MT to roughly 50--80\% for most NLU models.}
    \label{fig:model_comparison}
\end{figure}

~\section{Introduction}
Recent work has found that adaptive few-shot selection can improve the performance of LLMs on in-context learning tasks~\citep{agrawal2022incontextexamplesselectionmachine,chitale2024empiricalstudyincontextlearning, liu-etal-2022-makes}. 
In particular, several works adopt a ``retrieve-then-rerank'' approach, fetching a broad pool of candidates and using trained cross-encoders~\cite{li2023unifieddemonstrationretrieverincontext,wang2024learningretrieveincontextexamples,rubin-etal-2022-learning} or training-free scorers to select demonstrations for a specific inference sample~\cite{wu-etal-2023-self,peng2024revisitingdemonstrationselectionstrategies}. Separately, a line of work on adaptive/on-demand RAG uses model uncertainty or query complexity to decide \emph{whether} to retrieve or regenerate at all~\cite{jiang2023activeretrievalaugmentedgeneration,asai2023selfraglearningretrievegenerate,jeong2024adaptiveraglearningadaptretrievalaugmented,moskvoretskii-etal-2025-adaptive}, and other work validates or filters demonstrations after retrieval~\cite{zhang2025dvavalidatedemonstrationuse}. Our setting differs from both: ours is an in-context learning (ICL) demonstration-selection problem, distinct from RAG. Unlike open-ended RAG, we assume retrieval already happens over a fixed candidate pool, so our decision is not whether to retrieve but whether to \emph{rerank} the retrieved few-shot pool for in-context learning across MT and NLU tasks, using the same uncertainty-gating principle to decide when the extra reranking cost is worth paying. That cost is substantial: reranking can increase total token consumption by 13.4$\times$ for MT and 29.5$\times$ for NLU.\footnote{Token consumption is calculated per task as the total sum of tokens across all models and datasets.}

To address these concerns, we propose \emph{Training-Free Gated Reranking}, a simple yet effective approach, which uses the perplexity of the LLM's initial generation to rerank examples only when the model is uncertain. Rather than reranking every instance, we apply the reranking step exclusively when the input’s perplexity falls above a predefined threshold.

Our approach improves efficiency while matching performance across various models and datasets. Figure~\ref{fig:model_comparison} shows that tuning the perplexity threshold for reranking yields computational savings of 15\%-80\% while maintaining baseline performance. Our manual analysis confirms that reranking provides clear gains for difficult instances, but can degrade performance when applied to easy instances.
Overall, our approach presents a convenient tradeoff between resources and performance across 8 models ranging from 3B to 70B parameters, covering 18 combinations of language pairs and specialized domains, as well as 7 NLU datasets.\footnote{Code and experimental data are provided in the supplementary material for review, and will be made publicly available upon acceptance.}

\section{Perplexity-Based Reranking of Few-Shot Examples}

We propose a simple approach that adaptively allocates computational resources to choose few-shot examples based on the model's intrinsic uncertainty.

\paragraph{Formal definitions.}
Given an LLM $M$, an input $x$ and a large pool of candidate demonstrations $\mathcal{C} =((x_1, y_1), \ldots, (x_n, y_n))$, each consisting of an input and a corresponding gold label, ranked according to their relevance to $x$, we are looking for a function:
  
\begin{equation}
    T(M, x, \mathcal{C}) \mapsto \{c_1, \ldots, c_k | c_i \in \mathcal{C}\}
\end{equation}

That is, a selection of $k$ demonstration samples from the pool according to $x$.

\paragraph{Method.} 

We start by generating a prediction without reranking $\hat{y} = M(x, c_1, \ldots, c_k)$,  and use it to compute a normalized uncertainty score $U \in [0, 1]$. This score is derived from the inverse Conditional Perplexity of the generated target sequence.

\begin{equation}
\label{equ:uncertainty}
    \text{U}(M, x, \mathcal{C}) = 1 - \text{PPL}_M(\hat{y} \mid x, c_1, \ldots, c_k)^{-1}
\end{equation}

Where, and PPL computes the conditional perplexity of $\hat{y}$ on $x$ and the first $k$ examples.
$U$ is in the range $[0,1]$, where larger values indicate that the model is less certain in its own prediction.

We then define our demonstration selection function $T$ based on a predefined uncertainty threshold $\tau$:

\begin{equation}
    T(M, x, \mathcal{C})=
\begin{cases}
((x_1, y_1), \ldots (x_k, y_k))  & U \le \tau,\\
\operatorname{rerank}(M, x, \mathcal{C}) & U > \tau
\end{cases}
\end{equation}

where $rerank$ is defined as the conditional entropy rerank~\cite{peng2024revisitingdemonstrationselectionstrategies}:

\begin{equation}
\label{equ:condentrerank}
   \operatorname{rerank}(M, x, \mathcal{C}) = \operatorname*{Top-k}_{(x_i, y_i) \in \mathcal{C}} \Big( -\text{PPL}_M(x \mid (x_i, y_i)) \Big)
\end{equation}

In practice, we calibrate the threshold $\tau$ to maximize performance on a development set. During this calibration phase, we apply a moving average with a window size of 5 to smooth local variance.

\begin{table*}[tb!]
\centering
\resizebox{\textwidth}{!}{%
\begin{tabular}{l cc cc cc}
\toprule
 & \multicolumn{2}{c}{\textbf{Global (All)}} & \multicolumn{2}{c}{\textbf{Small Models}} & \multicolumn{2}{c}{\textbf{Bigger Models}} \\
\cmidrule(lr){2-3} \cmidrule(lr){4-5} \cmidrule(lr){6-7}
\textbf{Method} & \textbf{Perf. $\uparrow$} & \textbf{Save. \%} & \textbf{Perf. $\uparrow$} & \textbf{Save. \%} & \textbf{Perf. $\uparrow$} & \textbf{Save. \%} \\
\midrule
\multicolumn{7}{l}{\textbf{Machine Translation (BLEU / COMET)}} \\
\midrule
No Reranking & 37.09/81.05 & 100.0 & 33.37/79.46 & 100.0 & 38.33/81.57 & 100.0 \\
Full Reranking & 38.32/81.37 & 0.0 & 34.77/79.89 & 0.0 & 39.51/81.86 & 0.0 \\
Gating (dev-set calibrated) & \underline{38.42$^{\dagger}$}/\underline{81.39} & 20.86 {\scriptsize\textcolor{gray}{(33\% w.c.)}} & \underline{34.95$^{\dagger}$}/\underline{79.94} & 20.52 {\scriptsize\textcolor{gray}{(37\% w.c.)}} & \underline{39.58$^{\dagger}$}/\underline{81.88} & 20.98 {\scriptsize\textcolor{gray}{(22\% w.c.)}} \\
Gating (test-set calibrated) & \textbf{\underline{38.69$^{\dagger}$$^{\ddagger}$}}/\textbf{\underline{81.43}} & 17.42 {\scriptsize\textcolor{gray}{(27\% w.c.)}} & \textbf{\underline{35.19$^{\dagger}$$^{\ddagger}$}}/\textbf{\underline{79.98}} & 19.22 {\scriptsize\textcolor{gray}{(35\% w.c.)}} & \textbf{\underline{39.85$^{\dagger}$$^{\ddagger}$}}/\textbf{\underline{81.92}} & 16.82 {\scriptsize\textcolor{gray}{(2\% w.c.)}} \\
\midrule
\multicolumn{7}{l}{\textbf{Natural Language Understanding (Accuracy)}} \\
\midrule
No Reranking & 79.88 & 100.0 & 77.16 & 100.0 & 80.78 & 100.0 \\
Full Reranking & 80.73 & 0.0 & 78.45 & 0.0 & 81.50 & 0.0 \\
Gating (dev-set calibrated) & \underline{80.81} & 54.3 {\scriptsize\textcolor{gray}{(58\% w.c.)}} & \underline{78.60} & 50.2 {\scriptsize\textcolor{gray}{(61\% w.c.)}} & \underline{81.55} & 55.6 {\scriptsize\textcolor{gray}{(47\% w.c.)}} \\
Gating (test-set calibrated) & \textbf{\underline{81.41$^{\dagger}$}} & 47.16 {\scriptsize\textcolor{gray}{(24\% w.c.)}} & \textbf{\underline{79.30$^{\dagger}$}} & 46.69 {\scriptsize\textcolor{gray}{(25\% w.c.)}} & \textbf{\underline{82.11$^{\dagger}$$^{\ddagger}$}} & 47.32 {\scriptsize\textcolor{gray}{(23\% w.c.)}} \\
\bottomrule
\end{tabular}
   }
\caption{Performance summary across tasks and model sizes. Performance (Perf.) shows BLEU/COMET for MT and Accuracy for NLU. \textbf{Bold} indicates row maxima, and \underline{underlined} values exceed Full Reranking performance. $\tau$ denotes the gating threshold, calibrated on the dev or test set as indicated by each row. $^{\dagger}$ and $^{\ddagger}$ denote statistically significant improvements over No Reranking and Full Reranking, respectively ($p < 0.05$, two-sided). The grey parenthetical under each Save.\% column reports the mean measured wall-clock (w.c.) latency saving.}
\label{tab:final_merged_results}
\end{table*}

\section{Evaluation}
\subsection{Experimental Setup}
We list below key experimental details, see more details in the Appendix.

\paragraph{Tasks and models.}
We evaluate our approach across 8 LLMs spanning 3B--70B parameters, compared to a full reranking baseline~\cite{peng2024revisitingdemonstrationselectionstrategies}, assessing performance on both natural language understanding (NLU) and machine translation (MT) tasks. Specifically, we adopt the NLU benchmarks used by~\citet{peng2024revisitingdemonstrationselectionstrategies}: SST-2, SST-5~\cite{socher-etal-2013-recursive}, CR, AgNews, Subj~\cite{wang2019gluemultitaskbenchmarkanalysis}, MNLI~\cite{williams-etal-2018-broad}, and QNLI~\cite{wang2019gluemultitaskbenchmarkanalysis}. For MT, we use three domain specific corpora~\cite{koehn2017six}: EMEA (medical)~\cite{tiedemann-2012-parallel}, JRC-Acquis (legal)~\cite{steinberger-etal-2006-jrc}, and KDE (technical)~\cite{tiedemann-2012-parallel}. We consider both translation directions between English and Spanish, Portuguese, and German, yielding six directions per domain. We sample 1{,}000 parallel sentence pairs for each domain-direction combination, partitioned into 200 development examples for threshold tuning and 800 test examples for evaluation. All results are averaged over 20 random splits for statistical robustness.

\paragraph{Baselines.}
Baselines include retrieval-based $k$-shot selection with BM25~\cite{robertson2009probabilistic} and dense retrieval using e5-base-v2 or multilingual-e5-base depending on the language~\cite{wang2022text,wang2024multilinguale5textembeddings}, and always-rerank ($\tau=0$, reranking for every input at $O(N)$). We also compare gating signals to simple input proxies (e.g., source length, source entropy) and an uncertainty proxy (token-level logit gap - TARG)~\cite{wang2025targtrainingfreeadaptiveretrieval}.

\paragraph{Hyperparameters and metrics.}
\label{pa:hyperparams_and_metrics}
We fix $k=5$ demonstrations and a candidate pool of $N=100$, report MT quality with BLEU~\cite{papineni2002bleu} and COMET (wmt22-comet-da)~\cite{rei2022cometkiwi}, and NLU quality with accuracy. Following scaling laws for inference compute~\cite{kaplan2020scalinglawsneurallanguage} and standard open-weight serving metrics~\cite{griggs2024melangecostefficientlarge}, we evaluate efficiency using the unweighted sum of input and output tokens across all processing stages: (a) initial draft generation, (b) second-pass generation after reranking for triggered cases, and (c) all conditional perplexity computations over the $N=100$ candidates. We report relative savings against an always-reranking baseline. Stage (a) can be returned to the user immediately, matching no-reranking latency, the added wait from stage (b) is only incurred for inputs where the gate triggers reranking.

\subsection{Results}
Our experiments yield several interesting observations.

\paragraph{Always reranking is not the optimal policy for few-shot selection in MT and NLU.}
Applying reranking to every query is highly inefficient and unnecessary for maintaining quality. Table~\ref{tab:final_merged_results} demonstrates that our practical Empirical Gating method reduces average computational costs by 20.86\% and 54.3\% for MT and NLU, while achieving performance on par with Full Reranking (38.42 vs 38.32 avg BLEU, 80.81 vs 80.73 avg Accuracy). Furthermore, using a post-hoc test-set calibrated gating threshold achieves comparable cost reductions (17.42\% / 47.16\%) while slightly increasing Full Reranking performance (+0.37 avg BLEU, +0.68 avg Accuracy), demonstrating there is room for improvement.

\paragraph{Selective reranking via uncertainty can improve performance, especially for smaller models.}
Selective reranking remains highly competitive with the Full Reranking baseline across model scales. Table~\ref{tab:final_merged_results} highlights that this dynamic is most pronounced for small models: Full Reranking performance (34.77 avg BLEU/78.45 avg Accuracy) is slightly improved upon by our retrospective Gating (test-set calibrated) (+0.42 avg BLEU / +0.85 avg Accuracy) and practical Gating (dev-set calibrated) results (+0.18 avg BLEU / +0.15 avg Accuracy) while yielding average token savings of 19.22\% / 46.69\% and 20.52\% / 50.2\%. Larger models in the optimal Gating (test-set calibrated) settings show smaller yet positive margins (+0.34 avg BLEU / +0.61 avg Accuracy), while the Gating (dev-set calibrated) yields strong savings (20.98\% and 55.6\%) and comparable performance (+0.07 avg BLEU / +0.05 avg Accuracy) demonstrating that selective application remains a highly efficient alternative even at scale. Despite similar token-based savings, the measured wall-clock benefit is smaller for larger models than for smaller ones under dev-calibrated gating (22\% vs.\ 37\% for MT, 47\% vs.\ 61\% for NLU), indicating that the practical latency advantage shrinks somewhat at scale.

\paragraph{Selective reranking via uncertainty provides an efficient way to optimize performance with minimal impact on task performance.} 
As Figure~\ref{fig:model_comparison} illustrates, adjusting the gating threshold provides a way to balance cost and quality. This tradeoff relies on the  correlation between the LLM's initial uncertainty and the utility of the reranker.  
This allows us to reduce token consumption from 15\% to 80\%.

\paragraph{Uncertainty gating outperforms other gating methods.}
Alternative gating signals fail to achieve the same balance of quality and efficiency. As shown in Table~\ref{tab:gate_metric_ablation}, where we compare our approach against several alternative baseline metrics used for the same gating purpose: the perplexity of the input text (Source Entropy), the count of words (Word Length), along with  TARG~\citep{wang2025targtrainingfreeadaptiveretrieval}.

\begin{table}[t]
\centering
\small
\begin{tabular}{llcc}
\toprule
\textbf{Strategy} & \textbf{Task} & \textbf{Score} & \textbf{Savings} \\
\midrule
No Reranking & MT & 37.09 & 100.0\% \\
 & NLU & 79.88 & 100.0\% \\
\addlinespace
Full Reranking & MT & 38.32 & 0.0\% \\
 & NLU & 80.73 & 0.0\% \\
\midrule
\addlinespace
TARG & MT & 38.15 & 8.6\% \\
 & NLU & 80.46 & 35.6\% \\
\addlinespace
Source Entropy & MT & 38.19 & 11.8\% \\
 & NLU & 80.46$^{\ddagger}$ & 42.2\% \\
\addlinespace
Word Length & MT & 38.11$^{\ddagger}$ & 17.1\% \\
 & NLU & 80.37$^{\ddagger}$ & 38.1\% \\
 \addlinespace
 \textbf{\condppls{} (Ours)} & MT & \textbf{38.42} & 20.9\% \\
 & NLU & \textbf{80.81} & 54.3\% \\

\bottomrule
\end{tabular}
\caption{Comparison of gating indicators averaged across all models for MT and NLU. Thresholds ($\tau$) for each strategy are calibrated on a development set and evaluated on a test set. \condppls{} achieves the best performance-efficiency balance. $^{\ddagger}$ denotes statistical significance vs.\ full reranking ($p<0.05$, two-sided) --- note that \condppls{} is statistically indistinguishable from Full Reranking despite far higher savings.}
\label{tab:gate_metric_ablation}
\end{table}

\section{Manual Qualitative Analysis}

\begin{table}[tb!]
\small
\centering
\small
\begin{tabular}{lccc}
\toprule
\textbf{Error Cause} & \textbf{MT} & \textbf{NLU} & \textbf{Overall} \\
\midrule
\multicolumn{4}{c}{\textbf{Reranking errors}} \\
\midrule
Structural Template & 21 & 23 & 44 \\
High Variance & 4 & 7 & 11 \\
Other & 5 & 0 & 5 \\
\midrule
\multicolumn{4}{c}{\textbf{No Reranking errors}} \\
\midrule
Domain/Term Shift & 15 & 8 & 23 \\
High Variance & 9 & 9 & 18 \\
Relational Mapping & 0 & 9 & 9 \\
Bleu Misleading & 4 & 0 & 4 \\
Other & 2 & 4 & 6 \\
\bottomrule
\end{tabular}
\caption{Distribution of error causes when reranking degrades performance compared to baseline vs. when reranking improved performance on translation and NLU tasks.}
\label{tab:reranking_data}
\end{table}

To understand the mechanics driving the performance trade-offs of uncertainty-based gating, we manually annotate and analyze instances with the highest and lowest uncertainty scores. Specifically, we examine some of the top 30 instances (highest uncertainty, reranking beneficial) and bottom 30 instances (lowest uncertainty, reranking degraded) across both Machine Translation (MT) and Natural Language Understanding (NLU) tasks to diagnose how the few-shot examples influence the model's generation. See Table \ref{tab:reranking_data} and Appendix Table~\ref{tab:reranking_comparison_instances}.

\paragraph{High uncertainty reveals weak baseline retrieval.} Under high uncertainty, the baseline frequently retrieves misleading demonstrations, such as out-of-domain examples in MT (15 instances) or missed Relational Mappings in NLU (9 instances). The reranker resolves these issues by correctly identifying the task's structural logic and domain constraints. By successfully retrieving examples that enforce low variance, specialized terminology or exact relationships (e.g., mapping a neighborhood to a borough), reranking effectively grounds the model and reduces hallucinations.

\paragraph{Low uncertainty indicates strong baseline retrieval.} For low-uncertainty instances, the baseline already retrieves highly relevant, format-preserving contexts. Here, reranking can actually degrade performance. The primary cause across both MT (21 instances) and NLU (23 instances) is the disruption of Structural Templates. While the baseline retrieves near-exact syntactic templates, the reranker often overrides them with examples that are topically relevant but structurally varied. Furthermore, reranking can introduce high-variance examples with divergent meanings or lexical distractions that confuse the model. A small subset of MT errors fell into the "BLEU Misleading" category, where the reranker's valid translation was unfairly penalized by the metric.

\paragraph{Uncertainty does not always track gating correctness.} Uncertainty measures how confident the model is, not whether the demonstration it relies on is reliable. In Cases 6--6b
(Table~\ref{tab:reranking_comparison_instances_nlu}), reranking picks a demonstration that looks similar on the surface but is wrong, so the model copies its error. In Cases 7--7b (Table~\ref{tab:reranking_comparison_instances_nlu}), the
baseline confidently gives a wrong answer by matching the wrong cue (a template tag, a near-duplicate article), and reranking only helps because it happens to find a better demonstration, and Uncertainty did not detect that the baseline was incorrect. These exceptions come from a separate, targeted search for counterexamples, not from the annotated instances above.

\section{Conclusion}
We show that reranking few-shot examples is not always beneficial and can even hurt performance. Instead, we introduce \emph{Training-Free Gated Reranking}, which selectively reranks when the model is uncertain, reducing computation by 15\%-80\% while matching the performance of full reranking. Our practical, dev-set calibrated setting alone already saves an average of 20.86\%/54.3\% (MT/NLU) in computation at no cost to quality. Overall, our findings suggest that targeted reranking can break the usual tradeoff between higher computational cost and better downstream quality.

\section{Limitations}
Our approach presents three primary limitations. First, the gating mechanism relies on an empirically calibrated threshold from a development set. Distribution shifts between development and test data can cause suboptimal thresholding. Second, calculating the uncertainty score requires generating an initial, non-reranked prediction. For tasks with long output sequences, the latency of this initial step may reduce the efficiency gained from skipping the reranking phase. Finally, the method requires access to token-level perplexity, limiting its use to models and APIs that expose logits.

\bibliography{latex/custom}

@inproceedings{yu2024rankragunifyingcontextranking,
    author = {Yue Yu and
Wei Ping and
Zihan Liu and
Boxin Wang and
Jiaxuan You and
Chao Zhang and
Mohammad Shoeybi and
Bryan Catanzaro},
    bibsource = {dblp computer science bibliography, https://dblp.org},
    booktitle = {Advances in Neural Information Processing Systems 38: Annual Conference
on Neural Information Processing Systems 2024, NeurIPS 2024, Vancouver,
BC, Canada, December 10 - 15, 2024},
    editor = {Amir Globersons and
Lester Mackey and
Danielle Belgrave and
Angela Fan and
Ulrich Paquet and
Jakub M. Tomczak and
Cheng Zhang},
    title = {RankRAG: Unifying Context Ranking with Retrieval-Augmented Generation
in LLMs},
    url = {http://papers.nips.cc/paper\_files/paper/2024/hash/db93ccb6cf392f352570dd5af0a223d3-Abstract-Conference.html},
    year = {2024}
}

@inproceedings{jeong2024adaptiveraglearningadaptretrievalaugmented,
    address = {Mexico City, Mexico},
    author = {Jeong, Soyeong  and
Baek, Jinheon  and
Cho, Sukmin  and
Hwang, Sung Ju  and
Park, Jong},
    booktitle = {Proceedings of the 2024 Conference of the North American Chapter of the Association for Computational Linguistics: Human Language Technologies (Volume 1: Long Papers)},
    editor = {Duh, Kevin  and
Gomez, Helena  and
Bethard, Steven},
    pages = {7036--7050},
    publisher = {Association for Computational Linguistics},
    title = {Adaptive-{RAG}: Learning to Adapt Retrieval-Augmented Large Language Models through Question Complexity},
    url = {https://aclanthology.org/2024.naacl-long.389},
    year = {2024}
}

@misc{jiang2023activeretrievalaugmentedgeneration,
      title={Active Retrieval Augmented Generation}, 
      author={Zhengbao Jiang and Frank F. Xu and Luyu Gao and Zhiqing Sun and Qian Liu and Jane Dwivedi-Yu and Yiming Yang and Jamie Callan and Graham Neubig},
      year={2023},
      eprint={2305.06983},
      archivePrefix={arXiv},
      primaryClass={cs.CL},
      url={https://arxiv.org/abs/2305.06983}, 
}

@misc{asai2023selfraglearningretrievegenerate,
      title={Self-RAG: Learning to Retrieve, Generate, and Critique through Self-Reflection}, 
      author={Akari Asai and Zeqiu Wu and Yizhong Wang and Avirup Sil and Hannaneh Hajishirzi},
      year={2023},
      eprint={2310.11511},
      archivePrefix={arXiv},
      primaryClass={cs.CL},
      url={https://arxiv.org/abs/2310.11511}, 
}

@inproceedings{moskvoretskii-etal-2025-adaptive,
    title = "Adaptive Retrieval Without Self-Knowledge? Bringing Uncertainty Back Home",
    author = "Moskvoretskii, Viktor  and
      Marina, Maria  and
      Salnikov, Mikhail  and
      Ivanov, Nikolay  and
      Pletenev, Sergey  and
      Galimzianova, Daria  and
      Krayko, Nikita  and
      Konovalov, Vasily  and
      Nikishina, Irina  and
      Panchenko, Alexander",
    editor = "Che, Wanxiang  and
      Nabende, Joyce  and
      Shutova, Ekaterina  and
      Pilehvar, Mohammad Taher",
    booktitle = "Proceedings of the 63rd Annual Meeting of the Association for Computational Linguistics (Volume 1: Long Papers)",
    month = jul,
    year = "2025",
    address = "Vienna, Austria",
    publisher = "Association for Computational Linguistics",
    url = "https://aclanthology.org/2025.acl-long.319/",
    doi = "10.18653/v1/2025.acl-long.319",
    pages = "6355--6384",
    ISBN = "979-8-89176-251-0"
}

@misc{zhang2025dvavalidatedemonstrationuse,
      title={D.Va: Validate Your Demonstration First Before You Use It}, 
      author={Qi Zhang and Zhiqing Xiao and Ruixuan Xiao and Lirong Gao and Junbo Zhao},
      year={2025},
      eprint={2502.13646},
      archivePrefix={arXiv},
      primaryClass={cs.CL},
      url={https://arxiv.org/abs/2502.13646}, 
}

@misc{wang2025targtrainingfreeadaptiveretrieval,
    author = {Yufeng Wang and Lu wei and Haibin Ling},
    journal = {ArXiv preprint},
    title = {TARG: Training-Free Adaptive Retrieval Gating for Efficient RAG},
    url = {https://arxiv.org/abs/2511.09803},
    volume = {abs/2511.09803},
    year = {2025}
}

@misc{peng2024revisitingdemonstrationselectionstrategies,
    author = {Keqin Peng and Liang Ding and Yancheng Yuan and Xuebo Liu and Min Zhang and Yuanxin Ouyang and Dacheng Tao},
    journal = {ArXiv preprint},
    title = {Revisiting Demonstration Selection Strategies in In-Context Learning},
    url = {https://arxiv.org/abs/2401.12087},
    volume = {abs/2401.12087},
    year = {2024}
}

@misc{kadavath2022languagemodelsmostlyknow,
    author = {Saurav Kadavath and Tom Conerly and Amanda Askell and Tom Henighan and Dawn Drain and Ethan Perez and Nicholas Schiefer and Zac Hatfield-Dodds and Nova DasSarma and Eli Tran-Johnson and Scott Johnston and Sheer El-Showk and Andy Jones and Nelson Elhage and Tristan Hume and Anna Chen and Yuntao Bai and Sam Bowman and Stanislav Fort and Deep Ganguli and Danny Hernandez and Josh Jacobson and Jackson Kernion and Shauna Kravec and Liane Lovitt and Kamal Ndousse and Catherine Olsson and Sam Ringer and Dario Amodei and Tom Brown and Jack Clark and Nicholas Joseph and Ben Mann and Sam McCandlish and Chris Olah and Jared Kaplan},
    journal = {ArXiv preprint},
    title = {Language Models (Mostly) Know What They Know},
    url = {https://arxiv.org/abs/2207.05221},
    volume = {abs/2207.05221},
    year = {2022}
}

@inproceedings{tiedemann-2012-parallel,
    address = {Istanbul, Turkey},
    author = {Tiedemann, J{\"o}rg},
    booktitle = {Proceedings of the Eighth International Conference on Language Resources and Evaluation ({LREC}'12)},
    editor = {Calzolari, Nicoletta  and
Choukri, Khalid  and
Declerck, Thierry  and
Do{\u{g}}an, Mehmet U{\u{g}}ur  and
Maegaard, Bente  and
Mariani, Joseph  and
Moreno, Asuncion  and
Odijk, Jan  and
Piperidis, Stelios},
    pages = {2214--2218},
    publisher = {European Language Resources Association (ELRA)},
    title = {Parallel Data, Tools and Interfaces in {OPUS}},
    url = {http://www.lrec-conf.org/proceedings/lrec2012/pdf/463_Paper.pdf},
    year = {2012}
}

@inproceedings{steinberger-etal-2006-jrc,
    address = {Genoa, Italy},
    author = {Steinberger, Ralf  and
Pouliquen, Bruno  and
Widiger, Anna  and
Ignat, Camelia  and
Erjavec, Toma{\v{z}}  and
Tufi{\c{s}}, Dan  and
Varga, D{\'a}niel},
    booktitle = {Proceedings of the Fifth International Conference on Language Resources and Evaluation ({LREC}{'}06)},
    editor = {Calzolari, Nicoletta  and
Choukri, Khalid  and
Gangemi, Aldo  and
Maegaard, Bente  and
Mariani, Joseph  and
Odijk, Jan  and
Tapias, Daniel},
    publisher = {European Language Resources Association (ELRA)},
    title = {The {JRC}-{A}cquis: A Multilingual Aligned Parallel Corpus with 20+ Languages},
    url = {http://www.lrec-conf.org/proceedings/lrec2006/pdf/340_pdf.pdf},
    year = {2006}
}

@inproceedings{papineni2002bleu,
    address = {Philadelphia, Pennsylvania, USA},
    author = {Papineni, Kishore  and
Roukos, Salim  and
Ward, Todd  and
Zhu, Wei-Jing},
    booktitle = {Proceedings of the 40th Annual Meeting of the Association for Computational Linguistics},
    doi = {10.3115/1073083.1073135},
    editor = {Isabelle, Pierre  and
Charniak, Eugene  and
Lin, Dekang},
    pages = {311--318},
    publisher = {Association for Computational Linguistics},
    title = {{B}leu: a Method for Automatic Evaluation of Machine Translation},
    url = {https://aclanthology.org/P02-1040},
    year = {2002}
}

@inproceedings{rei2022cometkiwi,
    address = {Abu Dhabi, United Arab Emirates (Hybrid)},
    author = {Rei, Ricardo  and
Treviso, Marcos  and
Guerreiro, Nuno M.  and
Zerva, Chrysoula  and
Farinha, Ana C  and
Maroti, Christine  and
C. de Souza, Jos{\'e} G.  and
Glushkova, Taisiya  and
Alves, Duarte  and
Coheur, Luisa  and
Lavie, Alon  and
Martins, Andr{\'e} F. T.},
    booktitle = {Proceedings of the Seventh Conference on Machine Translation (WMT)},
    editor = {Koehn, Philipp  and
Barrault, Lo{\"\i}c  and
Bojar, Ond{\v{r}}ej  and
Bougares, Fethi  and
Chatterjee, Rajen  and
Costa-juss{\`a}, Marta R.  and
Federmann, Christian  and
Fishel, Mark  and
Fraser, Alexander  and
Freitag, Markus  and
Graham, Yvette  and
Grundkiewicz, Roman  and
Guzman, Paco  and
Haddow, Barry  and
Huck, Matthias  and
Jimeno Yepes, Antonio  and
Kocmi, Tom  and
Martins, Andr{\'e}  and
Morishita, Makoto  and
Monz, Christof  and
Nagata, Masaaki  and
Nakazawa, Toshiaki  and
Negri, Matteo  and
N{\'e}v{\'e}ol, Aur{\'e}lie  and
Neves, Mariana  and
Popel, Martin  and
Turchi, Marco  and
Zampieri, Marcos},
    pages = {634--645},
    publisher = {Association for Computational Linguistics},
    title = {{C}omet{K}iwi: {IST}-Unbabel 2022 Submission for the Quality Estimation Shared Task},
    url = {https://aclanthology.org/2022.wmt-1.60},
    year = {2022}
}

@article{wang2022text,
    author = {Liang Wang and Nan Yang and Xiaolong Huang and Binxing Jiao and Linjun Yang and Daxin Jiang and Rangan Majumder and Furu Wei},
    journal = {ArXiv preprint},
    title = {Text Embeddings by Weakly-Supervised Contrastive Pre-training},
    url = {https://arxiv.org/abs/2212.03533},
    volume = {abs/2212.03533},
    year = {2022}
}

@article{robertson2009probabilistic,
    author = {Stephen E. Robertson and Hugo Zaragoza},
    journal = {Found. Trends Inf. Retr.},
    pages = {333-389},
    title = {The Probabilistic Relevance Framework: BM25 and Beyond},
    url = {https://api.semanticscholar.org/CorpusID:207178704},
    volume = {3},
    year = {2009}
}

@inproceedings{koehn2017six,
    address = {Vancouver},
    author = {Koehn, Philipp  and
Knowles, Rebecca},
    booktitle = {Proceedings of the First Workshop on Neural Machine Translation},
    doi = {10.18653/v1/W17-3204},
    editor = {Luong, Thang  and
Birch, Alexandra  and
Neubig, Graham  and
Finch, Andrew},
    pages = {28--39},
    publisher = {Association for Computational Linguistics},
    title = {Six Challenges for Neural Machine Translation},
    url = {https://aclanthology.org/W17-3204},
    year = {2017}
}

@misc{wang2024multilinguale5textembeddings,
    author = {Liang Wang and Nan Yang and Xiaolong Huang and Linjun Yang and Rangan Majumder and Furu Wei},
    journal = {ArXiv preprint},
    title = {Multilingual E5 Text Embeddings: A Technical Report},
    url = {https://arxiv.org/abs/2402.05672},
    volume = {abs/2402.05672},
    year = {2024}
}

@misc{chitale2024empiricalstudyincontextlearning,
    author = {Pranjal A. Chitale and Jay Gala and Raj Dabre},
    journal = {ArXiv preprint},
    title = {An Empirical Study of In-context Learning in LLMs for Machine Translation},
    url = {https://arxiv.org/abs/2401.12097},
    volume = {abs/2401.12097},
    year = {2024}
}

@inproceedings{agrawal2022incontextexamplesselectionmachine,
    address = {Toronto, Canada},
    author = {Agrawal, Sweta  and
Zhou, Chunting  and
Lewis, Mike  and
Zettlemoyer, Luke  and
Ghazvininejad, Marjan},
    booktitle = {Findings of the Association for Computational Linguistics: ACL 2023},
    doi = {10.18653/v1/2023.findings-acl.564},
    editor = {Rogers, Anna  and
Boyd-Graber, Jordan  and
Okazaki, Naoaki},
    pages = {8857--8873},
    publisher = {Association for Computational Linguistics},
    title = {In-context Examples Selection for Machine Translation},
    url = {https://aclanthology.org/2023.findings-acl.564},
    year = {2023}
}

@misc{kaplan2020scalinglawsneurallanguage,
    author = {Jared Kaplan and Sam McCandlish and Tom Henighan and Tom B. Brown and Benjamin Chess and Rewon Child and Scott Gray and Alec Radford and Jeffrey Wu and Dario Amodei},
    journal = {ArXiv preprint},
    title = {Scaling Laws for Neural Language Models},
    url = {https://arxiv.org/abs/2001.08361},
    volume = {abs/2001.08361},
    year = {2020}
}

@inproceedings{wang2024learningretrieveincontextexamples,
    address = {St. Julian{'}s, Malta},
    author = {Wang, Liang  and
Yang, Nan  and
Wei, Furu},
    booktitle = {Proceedings of the 18th Conference of the European Chapter of the Association for Computational Linguistics (Volume 1: Long Papers)},
    editor = {Graham, Yvette  and
Purver, Matthew},
    pages = {1752--1767},
    publisher = {Association for Computational Linguistics},
    title = {Learning to Retrieve In-Context Examples for Large Language Models},
    url = {https://aclanthology.org/2024.eacl-long.105},
    year = {2024}
}

@inproceedings{li2023unifieddemonstrationretrieverincontext,
    address = {Toronto, Canada},
    author = {Li, Xiaonan  and
Lv, Kai  and
Yan, Hang  and
Lin, Tianyang  and
Zhu, Wei  and
Ni, Yuan  and
Xie, Guotong  and
Wang, Xiaoling  and
Qiu, Xipeng},
    booktitle = {Proceedings of the 61st Annual Meeting of the Association for Computational Linguistics (Volume 1: Long Papers)},
    doi = {10.18653/v1/2023.acl-long.256},
    editor = {Rogers, Anna  and
Boyd-Graber, Jordan  and
Okazaki, Naoaki},
    pages = {4644--4668},
    publisher = {Association for Computational Linguistics},
    title = {Unified Demonstration Retriever for In-Context Learning},
    url = {https://aclanthology.org/2023.acl-long.256},
    year = {2023}
}

@inproceedings{socher-etal-2013-recursive,
    address = {Seattle, Washington, USA},
    author = {Socher, Richard  and
Perelygin, Alex  and
Wu, Jean  and
Chuang, Jason  and
Manning, Christopher D.  and
Ng, Andrew  and
Potts, Christopher},
    booktitle = {Proceedings of the 2013 Conference on Empirical Methods in Natural Language Processing},
    editor = {Yarowsky, David  and
Baldwin, Timothy  and
Korhonen, Anna  and
Livescu, Karen  and
Bethard, Steven},
    pages = {1631--1642},
    publisher = {Association for Computational Linguistics},
    title = {Recursive Deep Models for Semantic Compositionality Over a Sentiment Treebank},
    url = {https://aclanthology.org/D13-1170},
    year = {2013}
}

@inproceedings{wang2019gluemultitaskbenchmarkanalysis,
    author = {Alex Wang and
Amanpreet Singh and
Julian Michael and
Felix Hill and
Omer Levy and
Samuel R. Bowman},
    bibsource = {dblp computer science bibliography, https://dblp.org},
    booktitle = {7th International Conference on Learning Representations, {ICLR} 2019,
New Orleans, LA, USA, May 6-9, 2019},
    publisher = {OpenReview.net},
    title = {{GLUE:} {A} Multi-Task Benchmark and Analysis Platform for Natural
Language Understanding},
    url = {https://openreview.net/forum?id=rJ4km2R5t7},
    year = {2019}
}

@inproceedings{williams-etal-2018-broad,
    address = {New Orleans, Louisiana},
    author = {Williams, Adina  and
Nangia, Nikita  and
Bowman, Samuel},
    booktitle = {Proceedings of the 2018 Conference of the North {A}merican Chapter of the Association for Computational Linguistics: Human Language Technologies, Volume 1 (Long Papers)},
    doi = {10.18653/v1/N18-1101},
    editor = {Walker, Marilyn  and
Ji, Heng  and
Stent, Amanda},
    pages = {1112--1122},
    publisher = {Association for Computational Linguistics},
    title = {A Broad-Coverage Challenge Corpus for Sentence Understanding through Inference},
    url = {https://aclanthology.org/N18-1101},
    year = {2018}
}

@misc{griggs2024melangecostefficientlarge,
    author = {Tyler Griggs and Xiaoxuan Liu and Jiaxiang Yu and Doyoung Kim and Wei-Lin Chiang and Alvin Cheung and Ion Stoica},
    journal = {ArXiv preprint},
    title = {M\'elange: Cost Efficient Large Language Model Serving by Exploiting GPU Heterogeneity},
    url = {https://arxiv.org/abs/2404.14527},
    volume = {abs/2404.14527},
    year = {2024}
}

@inproceedings{liu-etal-2022-makes,
    address = {Dublin, Ireland and Online},
    author = {Liu, Jiachang  and
Shen, Dinghan  and
Zhang, Yizhe  and
Dolan, Bill  and
Carin, Lawrence  and
Chen, Weizhu},
    booktitle = {Proceedings of Deep Learning Inside Out (DeeLIO 2022): The 3rd Workshop on Knowledge Extraction and Integration for Deep Learning Architectures},
    doi = {10.18653/v1/2022.deelio-1.10},
    editor = {Agirre, Eneko  and
Apidianaki, Marianna  and
Vuli{\'c}, Ivan},
    pages = {100--114},
    publisher = {Association for Computational Linguistics},
    title = {What Makes Good In-Context Examples for {GPT}-3?},
    url = {https://aclanthology.org/2022.deelio-1.10},
    year = {2022}
}

@inproceedings{rubin-etal-2022-learning,
    address = {Seattle, United States},
    author = {Rubin, Ohad  and
Herzig, Jonathan  and
Berant, Jonathan},
    booktitle = {Proceedings of the 2022 Conference of the North American Chapter of the Association for Computational Linguistics: Human Language Technologies},
    doi = {10.18653/v1/2022.naacl-main.191},
    editor = {Carpuat, Marine  and
de Marneffe, Marie-Catherine  and
Meza Ruiz, Ivan Vladimir},
    pages = {2655--2671},
    publisher = {Association for Computational Linguistics},
    title = {Learning To Retrieve Prompts for In-Context Learning},
    url = {https://aclanthology.org/2022.naacl-main.191},
    year = {2022}
}

@inproceedings{wu-etal-2023-self,
    address = {Toronto, Canada},
    author = {Wu, Zhiyong  and
Wang, Yaoxiang  and
Ye, Jiacheng  and
Kong, Lingpeng},
    booktitle = {Proceedings of the 61st Annual Meeting of the Association for Computational Linguistics (Volume 1: Long Papers)},
    doi = {10.18653/v1/2023.acl-long.79},
    editor = {Rogers, Anna  and
Boyd-Graber, Jordan  and
Okazaki, Naoaki},
    pages = {1423--1436},
    publisher = {Association for Computational Linguistics},
    title = {Self-Adaptive In-Context Learning: An Information Compression Perspective for In-Context Example Selection and Ordering},
    url = {https://aclanthology.org/2023.acl-long.79},
    year = {2023}
}

\section{Appendix}

\subsection{Prompt Template}
\label{app:prompts}

We utilized the following chat template for all Training-Free Gated Reranking experiments. Dynamic fields are denoted in brackets.

\begin{tcolorbox}[colback=gray!10, colframe=gray!60, title=\textbf{MT Prompt Template}]
\textbf{[System Message]} \\
You are an expert translator. Task: Translate [Source Language] to [Target Language].\\
Rule 1: Output ONLY the translated text.\\
Rule 2: Do not engage in conversation or explain the translation.\\
Rule 3: CRITICAL: You have been provided with a "Reference Translation" in the context. You MUST copy the terminology and structure from the Reference Translation exactly. Do not rephrase.

\textbf{[User Message]} \\
\#\#\# Reference Examples:

Input ([Source Language]): [Retrieved Source Text 1] \\
Output ([Target Language]): [Retrieved Target Text 1]

Input ([Source Language]): [Retrieved Source Text 2] \\
Output ([Target Language]): [Retrieved Target Text 2]

...

\#\#\# Current Task: \\
Input ([Source Language]): [Input Source Text] \\
Output ([Target Language]):
\end{tcolorbox}

\begin{tcolorbox}[colback=gray!10, colframe=gray!60, title=\textbf{NLU Prompt Template}]
\textbf{[System Message]} \\
You are an AI assistant. Complete the given task by outputting ONLY the required label. Do not explain.

\textbf{[User Message]} \\
\#\#\# Reference Examples:

[Retrieved Source Text 1] [Retrieved Target Label 1]

[Retrieved Source Text 2] [Retrieved Target Label 2]

...

\#\#\# Current Task: \\ {}
[Input Source Text]
\end{tcolorbox}

\subsection{Hyperparameter Selection}
To select our architectural choices, we conducted a three-stage ablation study on the KDE dataset using the \texttt{\llamass{}} model.

\paragraph{Retrieval Mechanism and Shot Count.}
First, we determined the optimal retrieval backbone. As shown in Table~\ref{tab:k_shot_ablation_v2}, \textbf{Dense-based retrieval} combined with reranking consistently outperforms lexical (BM25) and random selection. Crucially, the performance gap widens as the number of shots ($k$) increases, peaking at $k=5$. This validates our decision to fix $k=5$ for the main experiments, as it maximizes the context window's utility.

\begin{table}[ht]
\centering
\resizebox{\columnwidth}{!}{%
\begin{tabular}{l c ccc}
\toprule
\textbf{Strategy} & \textbf{$k$} & \textbf{COMET} & \textbf{BLEU} & \textbf{ChrF} \\
\midrule
\textbf{Zero-Shot} & 0 & 69.8 & 17.17 & 44.14 \\
\midrule
Random & 1 & \textbf{71.27} & 18.74 & 46.05 \\
                 & 3 & 71.26 & \textbf{19.16} & \textbf{46.07} \\
                 & 5 & 70.94 & 19.10 & 45.82 \\
\midrule
Dense         & 1 & 73.56 & 23.28 & 50.78 \\
                 & 3 & \textbf{73.98} & 24.35 & 51.80 \\
                 & 5 & 73.79 & \textbf{24.74} & \textbf{52.19} \\
\midrule
BM25& 1 & 73.09 & 23.86 & 50.65 \\
                 & 3 & 74.03 & \textbf{25.13} & 52.15 \\
                 & 5 & \textbf{74.29} & 24.97 & \textbf{52.33} \\
\midrule
BM25 Full Reranking & 1 & 73.85 & 24.40 & 51.38 \\
                 & 3 & 74.20 & 26.05 & 52.54 \\
                 & 5 & \textbf{74.51} & \textbf{26.21} & \textbf{52.81} \\
\midrule
Dense Full Reranking & 1 & 73.76 & 24.15 & 51.45 \\
                             & 3 & 74.65 & 26.30 & 53.19 \\
                             & 5 & \textbf{74.80} & \textbf{26.79} & \textbf{53.44} \\
\bottomrule
\end{tabular}%
   }
\caption{\textbf{Impact of Retrieval Strategy and Shot Count ($k$).} Ablation study on the KDE (En$\to$De) dataset. While static retrieval methods (Random, BM25, E5) show diminishing returns or saturation at higher $k$, the reranking strategies continue to improve, with \textbf{Dense Full Reranking \cite{peng2024revisitingdemonstrationselectionstrategies}} at $k=5$ establishing the upper bound for performance. Bold values indicate the optimal shot count for each specific strategy.}
\label{tab:k_shot_ablation_v2}
\end{table}

\paragraph{Comparison of Reranking Algorithms.}
To determine the optimal training-free method, we evaluated the effectiveness of \condpplrerank{} function against alternative strategies, including \srcentrerank{}, \logitgaprerank{} \cite{wang2025targtrainingfreeadaptiveretrieval}, and \rankragrerank{} \cite{yu2024rankragunifyingcontextranking}, which we apply without training. As shown in Table~\ref{tab:rerank_strategies}, while some reranking methods offer some gains, utilizing the \condpplrerank{} yields the most robust performance.

\begin{table}[ht]
    \centering
    \resizebox{\columnwidth}{!}{%
    \begin{tabular}{lccc}
        \toprule
        \textbf{Strategy} & \textbf{COMET} & \textbf{BLEU} & \textbf{ChrF} \\
        \midrule
        \srcentrerank{}   & 71.78 & 18.74 & 46.29 \\
        \rankragrerank{}  & 71.51 & 18.92 & 46.71 \\
        \logitgaprerank{} & 74.51 & 26.34 & 52.88 \\
        \condpplrerank{}  & \textbf{74.80} & \textbf{26.79} & \textbf{53.44} \\
        \bottomrule
    \end{tabular}%
    }
    \caption{\textbf{Re-Ranking Strategies.} Evaluation on KDE (En$\to$De) using \texttt{\llamas{}} (\llamass{}) ($k=5$).}
    \label{tab:rerank_strategies}
\end{table}

\paragraph{Sensitivity to Candidate Pool Size.}
Finally, we analyzed the computational trade-off regarding the cardinality of the candidate pool $|\mathcal{C}| = N$. As detailed in Table~\ref{tab:pool_size}, increasing the search space from $5 \times k$ to $20 \times k$ yields steady improvements. However, performance plateaus beyond $N=100$ ($20 \times k$). Consequently, we selected a pool size of 100 for all experiments to balance maximum utility with inference latency.

\begin{table}[ht]
    \footnotesize
    \centering
    \resizebox{\columnwidth}{!}{%
    \begin{tabular}{l ccc ccc}
        \toprule
        & \multicolumn{3}{c}{\textbf{Dense Full Reranking}} \\
        \cmidrule(lr){2-4} \cmidrule(lr){5-7}
        \textbf{MULT} & \textbf{COMET} & \textbf{BLEU} & \textbf{ChrF} \\
        \midrule
        1  & 74.04 & 24.95 & 52.42 \\
        5  & 74.43 & 26.01 & 52.80 \\
        10 & 74.42 & 26.20 & 53.08 \\
        20 & \textbf{74.88} & 26.80 & \textbf{53.47} \\
        30 & 74.83 & \textbf{27.08} & 53.38 \\
        \bottomrule
    \end{tabular}%
   }
    \caption{\textbf{Candidate Pool Sensitivity.} MULT is the pool-size multiplier ($N = \text{MULT} \times k$, with $k=5$); increasing the pool size yields gains up to $N=100$ (MULT=20), after which performance saturates.}
    \label{tab:pool_size}
\end{table}

\paragraph{Robustness to Simpler Prompts.} Our main MT system prompt instructs the model to copy terminology and structure from the retrieved reference translation exactly. To check that our gains are not specific to this prompt design, we reran the full gating pipeline for all models on all MT domain-directions with a simpler prompt, replacing the copy instruction with the plain system prompt \textit{"You are an AI assistant. Complete the given task by outputting ONLY the translated text. Do not explain."}, matching the style used for NLU. As shown in Table~\ref{tab:nocopy_ablation}, Gating (dev-set calibrated) remains competitive with Full Reranking under the simpler prompt, confirming the gating behavior is not specific to a particular prompt design

\begin{table}[ht]
    \centering
    \resizebox{\columnwidth}{!}{%
    \begin{tabular}{lcccc}
        \toprule
        \textbf{Model} & \textbf{Baseline} & \textbf{Full Rerank} & \textbf{Gated} & \textbf{Savings} \\
        \midrule
        Llama-3.2-3B-Instruct & 38.66 & 40.19 & \textbf{40.37} & 21.8\% \\
        Llama-3.3-70B-Instruct & 45.28 & 46.26 & \textbf{46.45} & 25.8\% \\
        Meta-Llama-3.1-8B-Instruct & 41.30 & 43.02 & \textbf{43.28} & 24.6\% \\
        Mistral-7B-Instruct-v0.3 & 39.21 & \textbf{40.74} & 40.71 & 15.5\% \\
        Qwen2.5-32B-Instruct & 43.20 & 44.42 & \textbf{44.51} & 24.5\% \\
        Qwen2.5-3B-Instruct & 35.37 & 36.47 & \textbf{36.82} & 24.9\% \\
        c4ai-command-r7b-12-2024 & 41.21 & 42.20 & \textbf{42.49} & 23.1\% \\
        aya-expanse-8b & 41.09 & \textbf{42.25} & 42.18 & 16.2\% \\
        \midrule
        Mean & 40.66 & 41.94 & \textbf{42.10} & 22.1\% \\
        \bottomrule
    \end{tabular}%
    }
    \caption{\textbf{No-Copy-Terminology Prompt Ablation (MT, BLEU).} Same gating pipeline as the main experiments, but with the copy-terminology/structure instruction removed from the system prompt. Gating uses dev-set calibrated thresholds; savings are token savings vs.\ Full Reranking. Bold marks the row maximum.}
    \label{tab:nocopy_ablation}
\end{table}

\paragraph{Alternative Gating Signal: Token-Level Self-Evaluation.}
As a further gating-signal ablation, we compare our \condppls{} gating signal against a token-level self-evaluation signal in the style of \citet{kadavath2022languagemodelsmostlyknow}, where the model judges the correctness of its own initial (unreranked) prediction and reranking is triggered when this self-judgment is negative. Table~\ref{tab:self_eval_ablation} reports this comparison for the same six models (3B--8B) on both MT and NLU. \condppls{} matches or exceeds Full Reranking performance at comparable or better savings, while self-evaluation gating trails Full Reranking performance for most models despite triggering reranking on a larger or smaller share of instances depending on the model, indicating our uncertainty signal is a more reliable gating trigger than self-evaluation at this model scale.

\begin{table}[t]
    \centering
    \footnotesize
    \setlength{\tabcolsep}{3pt}
    \begin{tabular}{lccc}
        \toprule
        \textbf{Model} & \textbf{Full RR} & \textbf{\condppls{}} & \textbf{Self-Eval} \\
        \midrule
        \multicolumn{4}{c}{\textit{Machine Translation (BLEU)}} \\
        \midrule
        \llamass{} & 35.69 & \textbf{35.73}$_{20\%}$ & 35.47$_{8\%}$ \\
        \qwenss{} & 33.86 & \textbf{34.16}$_{21\%}$ & 33.61$_{11\%}$ \\
        \llamams{} & 38.57 & \textbf{38.73}$_{26\%}$ & 38.47$_{14\%}$ \\
        \mistrals{} & 36.23 & \textbf{36.31}$_{15\%}$ & 35.55$_{57\%}$ \\
        \crbs{} & 36.96 & \textbf{37.16}$_{21\%}$ & 36.70$_{10\%}$ \\
        \ayas{} & 38.59 & \textbf{38.60}$_{19\%}$ & 35.71$_{84\%}$ \\
        \midrule
        \multicolumn{4}{c}{\textit{Natural Language Understanding (Accuracy)}} \\
        \midrule
        \llamass{} & 77.96 & \textbf{77.70}$_{46\%}$ & 77.14$_{34\%}$ \\
        \qwenss{} & 78.90 & \textbf{79.49}$_{55\%}$ & 78.63$_{44\%}$ \\
        \llamams{} & 79.64 & 79.60$_{55\%}$ & \textbf{80.01}$_{27\%}$ \\
        \mistrals{} & 78.51 & \textbf{79.30}$_{68\%}$ & 78.77$_{58\%}$ \\
        \crbs{} & 78.24 & 78.14$_{60\%}$ & \textbf{78.42}$_{36\%}$ \\
        \ayas{} & 80.72 & \textbf{80.78}$_{49\%}$ & 80.85$_{51\%}$ \\
        \bottomrule
    \end{tabular}
    \caption{\textbf{Self-Evaluation Gating Ablation.} \condppls{} vs.\ a token-level self-evaluation gating signal, both dev-set calibrated. Subscripts report token savings (\%) vs.\ Full Reranking (RR). Bold marks the higher-scoring gated variant per row.}
    \label{tab:self_eval_ablation}
\end{table}

\subsection{Detailed examples used in Analysis}
Table ~\ref{tab:reranking_comparison_instances} shows full details for the examples shown in the analysis section.

\begin{table*}[t]
\centering
\small
\resizebox{\textwidth}{!}{%
\setlength{\tabcolsep}{2pt}
\renewcommand{\arraystretch}{0.6}
\begin{tabularx}{\textwidth}{@{} >{\bfseries}p{0.06\textwidth} >{\raggedright\arraybackslash}X >{\raggedright\arraybackslash}X @{}}
\toprule
Strategy & Retrieved In-Context Example (Shot) & Final Model Output \\
\midrule

\multicolumn{3}{@{}l@{}}{\textbf{Case 1: Reranking Helped (Resolves Terminology Mismatch)}} \\
\multicolumn{3}{@{}p{\textwidth}@{}}{\textbf{Input (PT):} As actualizações sobre o decurso do estudo e sobre o recrutamento serão apresentadas juntamente com os reavaliação anual/ RPS. \textbf{Reference (EN):} Progress and recruitment updates will be provided alongside the Annual Re-assessment/ \textbf{PSURs}.} \\
\midrule
Baseline & 
\textit{Src:} As actualizações do progresso e dos recrutamentos serão apresentadas juntamente com os PSURs normais. \newline
\textit{Tgt:} Progress and recruitment updates will be provided alongside the standard PSURs. & 
Updates on the study course and recruitment will be presented together with the annual review/\textbf{RPS}. \\
\addlinespace
Rerank & 
\textit{Src:} O sumário da informação de segurança recolhida será apresentado juntamente com os reavaliação anual/ RPSs. \newline
\textit{Tgt:} Summary safety information gathered will be provided alongside the Annual Re-assessment/ PSURs. & 
Updates regarding the course of the study and recruitment will be provided alongside the Annual Re-assessment/ \textbf{PSURs}. \\
\midrule

\multicolumn{3}{@{}l@{}}{\textbf{Case 2: Baseline Was Better (Reranking Caused Hallucination)}} \\
\multicolumn{3}{@{}p{\textwidth}@{}}{\textbf{Input (EN):} Like all medicines, Xenical can cause side effects, although not everybody gets them. \textbf{Reference (PT):} \textbf{Como todos os medicamentos, Xenical pode causar efeitos secundários, no entanto estes não se manifestam em todas as pessoas.}} \\
\midrule
Baseline & 
\textit{Src:} Like all medicines, Arixtra can cause side effects, although not everybody gets them. \newline
\textit{Tgt:} Como todos os medicamentos, Arixtra pode causar efeitos secundários, no entanto, estes não se manifestam em todas as pessoas. & 
\textbf{Como todos os medicamentos, Xenical pode causar efeitos secundários, no entanto, estes não se manifestam em todas as pessoas.} \newline
\textcolor{gray}{\textit{Eng: Like all medicines, Xenical can cause side effects although not everybody gets them.}} \\
\addlinespace
Rerank & 
\textit{Src:} Take special care with XENICAL Weight loss may also affect the dose of medicines taken for other conditions... \newline
\textit{Tgt:} Tome especial cuidado com XENICAL A perda de peso pode também obrigar a alterar a dose de medicamentos... & 
\textbf{Tome especial cuidado com XENICAL A perda de peso pode também obrigar a alterar a dose de medicamentos que esteja a tomar para outras doenças (p. ex. colesterol elevado ou diabetes).} \newline
\textcolor{gray}{\textit{Eng: Take special care with XENICAL Weight loss may also affect the dose of medicines taken for other conditions (e.g. high cholesterol or diabetes).}} \\
\midrule

\multicolumn{3}{@{}l@{}}{\textbf{Case 3: Baseline Was Better (Reranking Caused Template Mismatch)}} \\
\multicolumn{3}{@{}p{\textwidth}@{}}{\textbf{Input (EN):} Like all medicines, STOCRIN can cause side effects, although not everybody gets them. \textbf{Reference (PT):} Como todos os medicamentos, STOCRIN pode causar \textbf{efeitos secundários, no entanto estes não se manifestam em todas as pessoas.}} \\
\midrule
Baseline & 
\textit{Src:} Like all medicines, Cetrotide can cause side effects, although not everybody gets them. \newline
\textit{Tgt:} Como os demais medicamentos, Cetrotide pode causar efeitos secundários, no entanto estes não se manifestam em todas as pessoas. & 
Como os demais medicamentos, STOCRIN pode causar \textbf{efeitos secundários, no entanto estes não se manifestam em todas as pessoas.} \newline
\textcolor{gray}{\textit{Eng: Like the other medicines, STOCRIN can cause \textbf{secondary effects, however these do not manifest in all people.}}} \\
\addlinespace
Rerank & 
\textit{Src:} Always notify your doctor immediately if you have these symptoms or any side effects while taking STOCRIN. \newline
\textit{Tgt:} Informe sempre o seu médico imediatamente se sentir qualquer um destes sintomas ou quaisquer efeitos indesejáveis... & 
Como todos os medicamentos, STOCRIN pode causar \textbf{efeitos colaterais, embora nem todo mundo os obtenha.} \newline
\textcolor{gray}{\textit{Eng: Like all medicines, STOCRIN can cause \textbf{collateral effects, although not everyone obtains them.}}} \\
\midrule

\multicolumn{3}{@{}l@{}}{\textbf{Case 4: Baseline Was Better (Reranking Lost Structural Template)}} \\
\multicolumn{3}{@{}p{\textwidth}@{}}{\textbf{Input Task (NLU):} Some rooms have balconies. Can we know Some rooms have balconies off of them that overlook the ocean.? \textbf{Reference Label:} \textbf{Maybe.}} \\
\midrule
Baseline & 
\textit{Task:} There are sea views from some rooms and there is also a swimming pool. Can we know Half the rooms are oceanview.? \newline
\textit{Label:} Maybe. & 
\textbf{Maybe.} \\
\addlinespace
Rerank & 
\textit{Task:} Rooms have private lanais, refrigerators, and coffeemakers. Can we know Rooms have refigerators.? \newline
\textit{Label:} Yes. & 
No. \\
\midrule

\multicolumn{3}{@{}l@{}}{\textbf{Case 5: Reranking Helped (Resolves Relational Mapping)}} \\
\multicolumn{3}{@{}p{\textwidth}@{}}{\textbf{Input Task (NLU):} In what borough is there a neighborhood called Huguenot? Can we know A small group of Huguenots also settled on the south shore of Staten Island along the New York Harbor, for which the current neighborhood of Huguenot was named.? \textbf{Reference Label:} \textbf{Yes.}} \\
\midrule
Baseline & 
\textit{Task:} The borough of Staten Island is primarily located on what island? Can we know This situation of boroughs separated by water led to the development of an extensive infrastructure of bridges and tunnels.? \newline
\textit{Label:} No. & 
No. \\
\addlinespace
Rerank & 
\textit{Task:} In what borough is the Douglaston neighborhood located? Can we know In contrast, New York City also has neighborhoods that are less densely populated and feature free-standing dwellings.? \newline
\textit{Label:} No. &
\textbf{Yes.} \\
\bottomrule
\end{tabularx}%
   }
\caption{Comparison of Baseline vs. Reranking In-Context Retrieval on Translation and NLU Quality}
\label{tab:reranking_comparison_instances}
\end{table*}

 \begin{table*}[t]
  \centering
  \small
  \resizebox{\textwidth}{!}{%
  \setlength{\tabcolsep}{2pt}
  \renewcommand{\arraystretch}{0.6}
  \begin{tabularx}{\textwidth}{@{} >{\bfseries}p{0.06\textwidth} >{\raggedright\arraybackslash}X >{\raggedright\arraybackslash}X @{}}
  \toprule
  Strategy & Retrieved In-Context Example (Shot) & Final Model Output \\
  \midrule

  \multicolumn{3}{@{}l@{}}{\textbf{Case 6: Reranking Hurt Despite High Uncertainty (Retrieved a Noisy Demonstration)}} \\
  \multicolumn{3}{@{}p{\textwidth}@{}}{\textbf{Model:} \mistrals{} \quad \textbf{Input (EN):} Need to reload medium to return to proper state. \textbf{Reference (PT):} É necessário carregar de novo o dispositivo para voltar a um estado adequado.
  \textbf{Uncertainty:} 0.81} \\
  \midrule
  Baseline &
  \textit{Src:} (Re)Load the medium \newline
  \textit{Tgt:} Reler o dispositivo &
  \textbf{Precisa se recarregar o dispositivo para retornar ao estado correto.} \newline
  \textcolor{gray}{\textit{Eng: [You] need to reload the device to return to the correct state.}} \\
  \addlinespace
  Rerank &
  \textit{Src:} (Re)Load the medium \newline
  \textit{Tgt:} Reler o dispositivo &
  Precisa se reler o meio para retornar ao estado correto. \newline
  \textcolor{gray}{\textit{Eng: [You] need to re-read the medium to return to the correct state.}} \\
  \midrule

  \multicolumn{3}{@{}l@{}}{\textbf{Case 6b: Reranking Hurt Despite High Uncertainty (Lexical-Overlap Trap, NLU)}} \\
  \multicolumn{3}{@{}p{\textwidth}@{}}{\textbf{Model:} \ayas{} \quad \textbf{Input Task (NLU):} Review: "good movie ." Sentiment: \textbf{Reference Label:} \textbf{great} \textbf{Uncertainty:} 0.92} \\
  \midrule
  Baseline &
  \textit{Task:} Review: "a compelling film." Sentiment: \newline
  \textit{Label:} great &
  \textbf{great} \\
  \addlinespace
  Rerank &
  \textit{Task:} Review: "nice piece of work." Sentiment: \newline
  \textit{Label:} good &
  good \\
  \midrule

  \multicolumn{3}{@{}l@{}}{\textbf{Case 7: Baseline Failed Despite Low Uncertainty (Confident Degenerate Output)}} \\
  \multicolumn{3}{@{}p{\textwidth}@{}}{\textbf{Model:} \crbs{} \quad \textbf{Input (EN):} Set here the name of content creator. \textbf{Reference (DE):} Geben Sie hier den Namen des Erstellers des Inhalts an. \textbf{Uncertainty:} 0.09} \\
  \midrule
  Baseline &
  \textit{Src:} Set here the original owner of content. \newline
  \textit{Tgt:} Geben Sie hier Originalbesitzer des Inhalts an. &
  Input (en): Set here the name of content creator. \newline
\textcolor{gray}{\textit{[No translation produced -- the model echoes the prompt's own "Input (en):" template tag instead of translating.]}} \\
  \addlinespace
  Rerank &
  \textit{Src:} Set here the original owner of content. This field is limited to 32 ASCII characters. \newline
  \textit{Tgt:} Geben Sie hier den Originalbesitzer des Inhalts an... &
  \textbf{Geben Sie hier den Namen des Inhaltserstellers ein.} \newline
  \textcolor{gray}{\textit{Eng: Enter the name of the content creator here.}} \\
  \midrule

  \multicolumn{3}{@{}l@{}}{\textbf{Case 7b: Baseline Failed Despite Low Uncertainty (Near-Duplicate-Article Trap, NLU)}} \\
  \multicolumn{3}{@{}p{\textwidth}@{}}{\textbf{Model:} \llamams{} \quad \textbf{Input Task (NLU):} Article: "Sprint To Spend \$3 Billion on Network Upgrades... wireless network, including... high-speed Internet services." Category:
  \textbf{Reference Label:} \textbf{Sci/Tech} \textbf{Uncertainty:} $<$0.01} \\
  \midrule
  Baseline &
  \textit{Task:} Article: "Sprint upgrades mobile network... spend \$3 billion... upgrade its mobile network including... high-speed Internet services." Category: \newline
  \textit{Label:} Business &
  Business \\
  \addlinespace
  Rerank &
  \textit{Task:} Article: "Wireless Revenues Top \$100 Billion, Says Census Bureau..." Category: \newline
  \textit{Label:} Sci/Tech &
  \textbf{Sci/Tech} \\
  \bottomrule
  \end{tabularx}%
     }
  \caption{Additional MT and NLU Case Examples for Baseline vs.\ Reranking Comparison (continued from Table~\ref{tab:reranking_comparison_instances}).}
  \label{tab:reranking_comparison_instances_nlu}
  \end{table*}

\subsection{Performance Across Tasks and Models}
Table \ref{tab:per_dataset_summary_combined} details the results across individual datasets, while Table \ref{tab:per_model_summary_combined} provides a model-by-model breakdown. Tables~\ref{tab:heatmap_results_test_calibrated} and \ref{tab:heatmap_results_dev_calibrated} show Gated (test-set calibrated) and Gated (dev-set calibrated) results per model and per dataset. The data indicate that our gating strategy generally yields higher performance than full reranking while simultaneously maintaining computational efficiency in the large majority of configurations. Although empirical performance occasionally trails that of full reranking, and a small number of configurations show negative token savings (see the discussion following Tables~\ref{tab:heatmap_results_dev_calibrated} and~\ref{tab:heatmap_results_test_calibrated}), efficiency gains hold for the vast majority of model-dataset combinations.

\begin{table*}[ht]
\centering
\resizebox{\textwidth}{!}{%
\begin{tabular}{lcccccc}
\toprule
\textbf{Dataset} & \textbf{Baseline} & \textbf{Full Rerank} & \textbf{Gating (test-set calibrated)} & Savings & \textbf{Gating (dev-set calibrated)} & Savings \\
\midrule
\multicolumn{7}{c}{\textbf{Machine Translation (BLEU)}} \\
\midrule
EMEA & 38.77 & 39.89 & \textbf{40.46}$^{\dagger}$$^{\ddagger}$ & 26\% {\scriptsize\textcolor{gray}{(32\% w.c.)}} & \textbf{40.12}$^{\dagger}$ & 29\% {\scriptsize\textcolor{gray}{(41\% w.c.)}} \\
JRC-Acquis & 39.39 & 40.83 & \textbf{41.07}$^{\dagger}$$^{\ddagger}$ & 12\% {\scriptsize\textcolor{gray}{(13\% w.c.)}} & \textbf{40.87}$^{\dagger}$ & 16\% {\scriptsize\textcolor{gray}{(21\% w.c.)}} \\
KDE & 33.11 & 34.26 & \textbf{34.53}$^{\dagger}$$^{\ddagger}$ & 15\% {\scriptsize\textcolor{gray}{(36\% w.c.)}} & \textbf{34.27}$^{\dagger}$ & 18\% {\scriptsize\textcolor{gray}{(42\% w.c.)}} \\
\midrule
\multicolumn{7}{c}{\textbf{Natural Language Understanding (Accuracy)}} \\
\midrule
AgNews & 91.66 & 92.63 & \textbf{92.67}$^{\dagger}$$^{\ddagger}$ & 47\% {\scriptsize\textcolor{gray}{(1\% w.c.)}} & 92.34$^{\dagger}$ & 58\% {\scriptsize\textcolor{gray}{(71\% w.c.)}} \\
CR & 93.76 & 93.62 & \textbf{94.61}$^{\dagger}$$^{\ddagger}$ & 88\% {\scriptsize\textcolor{gray}{(54\% w.c.)}} & \textbf{93.72} & 80\% {\scriptsize\textcolor{gray}{(82\% w.c.)}} \\
MNLI & 65.90 & 67.50 & \textbf{67.75}$^{\dagger}$$^{\ddagger}$ & 15\% {\scriptsize\textcolor{gray}{(13\% w.c.)}} & 67.13$^{\dagger}$ & 45\% {\scriptsize\textcolor{gray}{(4\% w.c.)}} \\
QNLI & 72.10 & 73.58 & \textbf{74.04}$^{\dagger}$ & 34\% {\scriptsize\textcolor{gray}{(8\% w.c.)}} & \textbf{73.63} & 47\% {\scriptsize\textcolor{gray}{(30\% w.c.)}} \\
SST-2 & 93.37 & 94.02 & \textbf{94.34}$^{\dagger}$ & 67\% {\scriptsize\textcolor{gray}{(47\% w.c.)}} & 93.89 & 59\% {\scriptsize\textcolor{gray}{(54\% w.c.)}} \\
SST-5 & 51.81 & 51.34 & \textbf{53.16}$^{\dagger}$$^{\ddagger}$ & 62\% {\scriptsize\textcolor{gray}{(47\% w.c.)}} & \textbf{52.25}$^{\ddagger}$ & 60\% {\scriptsize\textcolor{gray}{(72\% w.c.)}} \\
Subj & 90.56 & 92.46 & \textbf{93.27}$^{\dagger}$$^{\ddagger}$ & 17\% {\scriptsize\textcolor{gray}{(6\% w.c.)}} & \textbf{92.75}$^{\dagger}$$^{\ddagger}$ & 30\% {\scriptsize\textcolor{gray}{(21\% w.c.)}} \\
\bottomrule
\end{tabular}%
}
\caption{Performance breakdown by dataset using \condppls{}. Gated (test-set calibrated) represents the theoretical upper bound, while Gated (dev-set calibrated) represents practical levels. Gated values exceeding Full Reranking are in bold. $\dagger$ denotes statistical significance against the baseline. $^{\ddagger}$ sig.\ vs.\ full reranking ($p<0.05$, two-sided). Grey parentheticals report measured wall-clock (w.c.) latency savings, averaged over the 8 models for which timing was measured, under each column's own $\tau$ calibration; MT domain rows average over the six language directions within that domain.}
\label{tab:per_dataset_summary_combined}
\end{table*}

\begin{table*}[ht]
\centering
\resizebox{\textwidth}{!}{%
\begin{tabular}{lcccccc}
\toprule
\textbf{Model} & \textbf{Baseline} & \textbf{Full Rerank} & \textbf{Gated (Test-Set Calibrated)} & Savings & \textbf{Gated (Dev-Set Calibrated)} & Savings \\
\midrule
\multicolumn{7}{c}{\textbf{Machine Translation (BLEU)}} \\
\midrule
Llama-3.2-3B-Instruct & 34.22 & 35.69 & \textbf{36.00}$^{\dagger}$$^{\ddagger}$ & 19\% {\scriptsize\textcolor{gray}{(24\% w.c.)}} & \textbf{35.73}$^{\dagger}$ & 20\% {\scriptsize\textcolor{gray}{(27\% w.c.)}} \\
Llama-3.3-70B-Instruct & 43.16 & 44.16 & \textbf{44.38}$^{\dagger}$$^{\ddagger}$ & 18\% {\scriptsize\textcolor{gray}{(-3\% w.c.)}} & 44.16$^{\dagger}$ & 24\% {\scriptsize\textcolor{gray}{(18\% w.c.)}} \\
Meta-Llama-3.1-8B-Instruct & 37.28 & 38.57 & \textbf{39.03}$^{\dagger}$$^{\ddagger}$ & 22\% {\scriptsize\textcolor{gray}{(27\% w.c.)}} & \textbf{38.73}$^{\dagger}$ & 26\% {\scriptsize\textcolor{gray}{(36\% w.c.)}} \\
Mistral-7B-Instruct-v0.3 & 35.00 & 36.23 & \textbf{36.53}$^{\dagger}$ & 11\% {\scriptsize\textcolor{gray}{(26\% w.c.)}} & \textbf{36.31}$^{\dagger}$ & 15\% {\scriptsize\textcolor{gray}{(30\% w.c.)}} \\
Qwen2.5-32B-Instruct & 41.38 & 42.53 & \textbf{42.74}$^{\dagger}$$^{\ddagger}$ & 16\% {\scriptsize\textcolor{gray}{(7\% w.c.)}} & 42.52$^{\dagger}$ & 21\% {\scriptsize\textcolor{gray}{(26\% w.c.)}} \\
Qwen2.5-3B-Instruct & 32.53 & 33.86 & \textbf{34.38}$^{\dagger}$$^{\ddagger}$ & 19\% {\scriptsize\textcolor{gray}{(35\% w.c.)}} & \textbf{34.16}$^{\dagger}$ & 21\% {\scriptsize\textcolor{gray}{(36\% w.c.)}} \\
aya-expanse-8b & 37.64 & 38.59 & \textbf{38.84}$^{\dagger}$$^{\ddagger}$ & 14\% {\scriptsize\textcolor{gray}{(80\% w.c.)}} & \textbf{38.60}$^{\dagger}$ & 19\% {\scriptsize\textcolor{gray}{(80\% w.c.)}} \\
c4ai-command-r7b-12-2024 & 35.52 & 36.96 & \textbf{37.58}$^{\dagger}$$^{\ddagger}$ & 21\% {\scriptsize\textcolor{gray}{(16\% w.c.)}} & \textbf{37.16}$^{\dagger}$ & 21\% {\scriptsize\textcolor{gray}{(18\% w.c.)}} \\
\midrule
\multicolumn{7}{c}{\textbf{Natural Language Understanding (Accuracy)}} \\
\midrule
Llama-3.2-3B-Instruct & 76.32 & 77.99 & \textbf{78.47}$^{\dagger}$ & 36\% {\scriptsize\textcolor{gray}{(32\% w.c.)}} & 77.70$^{\dagger}$ & 46\% {\scriptsize\textcolor{gray}{(79\% w.c.)}} \\
Llama-3.3-70B-Instruct & 84.12 & 85.05 & \textbf{85.55}$^{\dagger}$$^{\ddagger}$ & 36\% {\scriptsize\textcolor{gray}{(7\% w.c.)}} & 84.93 & 48\% {\scriptsize\textcolor{gray}{(57\% w.c.)}} \\
Meta-Llama-3.1-8B-Instruct & 79.06 & 79.74 & \textbf{80.21}$^{\dagger}$$^{\ddagger}$ & 47\% {\scriptsize\textcolor{gray}{(34\% w.c.)}} & 79.60$^{\ddagger}$ & 55\% {\scriptsize\textcolor{gray}{(82\% w.c.)}} \\
Mistral-7B-Instruct-v0.3 & 78.31 & 78.55 & \textbf{79.76}$^{\dagger}$$^{\ddagger}$ & 62\% {\scriptsize\textcolor{gray}{(-3\% w.c.)}} & \textbf{79.30} & 68\% {\scriptsize\textcolor{gray}{(68\% w.c.)}} \\
Qwen2.5-32B-Instruct & 85.91 & 86.67 & \textbf{87.13}$^{\dagger}$ & 49\% {\scriptsize\textcolor{gray}{(48\% w.c.)}} & 86.59$^{\dagger}$ & 54\% {\scriptsize\textcolor{gray}{(36\% w.c.)}} \\
Qwen2.5-3B-Instruct & 78.01 & 78.91 & \textbf{80.14}$^{\dagger}$$^{\ddagger}$ & 57\% {\scriptsize\textcolor{gray}{(29\% w.c.)}} & \textbf{79.49} & 55\% {\scriptsize\textcolor{gray}{(61\% w.c.)}} \\
aya-expanse-8b & 80.06 & 80.74 & \textbf{81.24}$^{\dagger}$ & 47\% {\scriptsize\textcolor{gray}{(26\% w.c.)}} & \textbf{80.78} & 49\% {\scriptsize\textcolor{gray}{(10\% w.c.)}} \\
c4ai-command-r7b-12-2024 & 77.24 & 78.22 & \textbf{78.76}$^{\dagger}$$^{\ddagger}$ & 43\% {\scriptsize\textcolor{gray}{(33\% w.c.)}} & 78.14 & 60\% {\scriptsize\textcolor{gray}{(72\% w.c.)}} \\
\bottomrule
\end{tabular}%
}
\caption{Performance breakdown by model using \condppls{}. Gated (Test-Set Calibrated) represents the theoretical upper bound, while Gated (Dev-Set Calibrated) represents practical levels. Gated values exceeding Full Reranking are in bold. $\dagger$ denotes statistical significance against the baseline. $^{\ddagger}$ sig.\ vs.\ full reranking (two-sided, $p<0.05$). Grey parentheticals report measured wall-clock (w.c.) latency savings against full reranking, computed under each column's own $\tau$ calibration.}
\label{tab:per_model_summary_combined}
\end{table*}

\begin{table*}[t]
\centering
\resizebox{\textwidth}{!}{%
\begin{tabular}{l c c c c c c c c}
\hline
Dataset & Qwen 3B & Llama 3B & Command & Mistral & Aya & Llama 8B & Qwen 32B & Llama 70B \\
\hline
EMEA (de$\rightarrow$en) & \cellcolor[HTML]{E3F4DE} \begin{tabular}{@{}c@{}}+0.44 BLEU$^{\ddagger}$ \\ (13.4\%)\end{tabular} & \cellcolor[HTML]{D4EECE} \begin{tabular}{@{}c@{}}+0.18 BLEU$^{\ddagger}$ \\ (19.9\%)\end{tabular} & \cellcolor[HTML]{DBF1D5} \begin{tabular}{@{}c@{}}+0.76 BLEU$^{\ddagger}$ \\ (17.0\%)\end{tabular} & \cellcolor[HTML]{DAF0D4} \begin{tabular}{@{}c@{}}+0.21 BLEU$^{\ddagger}$ \\ (17.3\%)\end{tabular} & \cellcolor[HTML]{A2D99C} \begin{tabular}{@{}c@{}}-0.06 BLEU \\ (37.2\%)\end{tabular} & \cellcolor[HTML]{B1E0AB} \begin{tabular}{@{}c@{}}+0.25 BLEU$^{\ddagger}$ \\ (32.4\%)\end{tabular} & \cellcolor[HTML]{9FD899} \begin{tabular}{@{}c@{}}+0.12 BLEU \\ (38.2\%)\end{tabular} & \cellcolor[HTML]{C0E6B9} \begin{tabular}{@{}c@{}}-0.27 BLEU$^{\ddagger}$ \\ (27.7\%)\end{tabular} \\
EMEA (en$\rightarrow$de) & \cellcolor[HTML]{C3E7BC} \begin{tabular}{@{}c@{}}+0.41 BLEU$^{\ddagger}$ \\ (26.5\%)\end{tabular} & \cellcolor[HTML]{D6EFD0} \begin{tabular}{@{}c@{}}+0.41 BLEU$^{\ddagger}$ \\ (19.1\%)\end{tabular} & \cellcolor[HTML]{D8F0D2} \begin{tabular}{@{}c@{}}-3.47 BLEU$^{\ddagger}$ \\ (18.2\%)\end{tabular} & \cellcolor[HTML]{84CC83} \begin{tabular}{@{}c@{}}+0.26 BLEU$^{\ddagger}$ \\ (45.4\%)\end{tabular} & \cellcolor[HTML]{B5E1AE} \begin{tabular}{@{}c@{}}+0.53 BLEU$^{\ddagger}$ \\ (31.1\%)\end{tabular} & \cellcolor[HTML]{BEE5B8} \begin{tabular}{@{}c@{}}+0.70 BLEU$^{\ddagger}$ \\ (27.8\%)\end{tabular} & \cellcolor[HTML]{AADDA4} \begin{tabular}{@{}c@{}}-0.11 BLEU$^{\ddagger}$ \\ (34.5\%)\end{tabular} & \cellcolor[HTML]{A7DBA0} \begin{tabular}{@{}c@{}}-0.34 BLEU$^{\ddagger}$ \\ (35.9\%)\end{tabular} \\
EMEA (en$\rightarrow$es) & \cellcolor[HTML]{97D492} \begin{tabular}{@{}c@{}}+0.38 BLEU$^{\ddagger}$ \\ (40.4\%)\end{tabular} & \cellcolor[HTML]{A2D99C} \begin{tabular}{@{}c@{}}+0.12 BLEU$^{\ddagger}$ \\ (37.1\%)\end{tabular} & \cellcolor[HTML]{D9F0D3} \begin{tabular}{@{}c@{}}+0.23 BLEU$^{\ddagger}$ \\ (17.7\%)\end{tabular} & \cellcolor[HTML]{DEF2D9} \begin{tabular}{@{}c@{}}+0.08 BLEU \\ (15.5\%)\end{tabular} & \cellcolor[HTML]{84CC83} \begin{tabular}{@{}c@{}}+0.42 BLEU$^{\ddagger}$ \\ (45.7\%)\end{tabular} & \cellcolor[HTML]{5DB96B} \begin{tabular}{@{}c@{}}+0.70 BLEU$^{\ddagger}$ \\ (55.8\%)\end{tabular} & \cellcolor[HTML]{A7DBA0} \begin{tabular}{@{}c@{}}-0.12 BLEU$^{\ddagger}$ \\ (35.9\%)\end{tabular} & \cellcolor[HTML]{C2E7BB} \begin{tabular}{@{}c@{}}-0.03 BLEU \\ (26.6\%)\end{tabular} \\
EMEA (es$\rightarrow$en) & \cellcolor[HTML]{BAE3B3} \begin{tabular}{@{}c@{}}+0.50 BLEU$^{\ddagger}$ \\ (29.5\%)\end{tabular} & \cellcolor[HTML]{B8E3B2} \begin{tabular}{@{}c@{}}+0.36 BLEU$^{\ddagger}$ \\ (29.9\%)\end{tabular} & \cellcolor[HTML]{9ED798} \begin{tabular}{@{}c@{}}+1.17 BLEU$^{\ddagger}$ \\ (38.5\%)\end{tabular} & \cellcolor[HTML]{D7EFD1} \begin{tabular}{@{}c@{}}-0.20 BLEU$^{\ddagger}$ \\ (18.7\%)\end{tabular} & \cellcolor[HTML]{E5F5E1} \begin{tabular}{@{}c@{}}+0.36 BLEU$^{\ddagger}$ \\ (12.1\%)\end{tabular} & \cellcolor[HTML]{BAE3B3} \begin{tabular}{@{}c@{}}+0.28 BLEU$^{\ddagger}$ \\ (29.5\%)\end{tabular} & \cellcolor[HTML]{B1E0AB} \begin{tabular}{@{}c@{}}+0.03 BLEU \\ (32.1\%)\end{tabular} & \cellcolor[HTML]{CBEBC5} \begin{tabular}{@{}c@{}}-0.27 BLEU$^{\ddagger}$ \\ (23.1\%)\end{tabular} \\
EMEA (en$\rightarrow$pt) & \cellcolor[HTML]{C9EAC2} \begin{tabular}{@{}c@{}}+0.60 BLEU$^{\ddagger}$ \\ (24.2\%)\end{tabular} & \cellcolor[HTML]{D3EECD} \begin{tabular}{@{}c@{}}-0.03 BLEU \\ (20.1\%)\end{tabular} & \cellcolor[HTML]{B6E2AF} \begin{tabular}{@{}c@{}}+2.40 BLEU$^{\ddagger}$ \\ (30.9\%)\end{tabular} & \cellcolor[HTML]{2A924A}\color{white} \begin{tabular}{@{}c@{}}+1.77 BLEU$^{\ddagger}$ \\ (72.1\%)\end{tabular} & \cellcolor[HTML]{6ABF71} \begin{tabular}{@{}c@{}}+0.23 BLEU$^{\ddagger}$ \\ (52.7\%)\end{tabular} & \cellcolor[HTML]{91D28E} \begin{tabular}{@{}c@{}}+1.28 BLEU$^{\ddagger}$ \\ (41.9\%)\end{tabular} & \cellcolor[HTML]{C8E9C1} \begin{tabular}{@{}c@{}}+0.42 BLEU$^{\ddagger}$ \\ (24.8\%)\end{tabular} & \cellcolor[HTML]{A5DB9F} \begin{tabular}{@{}c@{}}+1.18 BLEU$^{\ddagger}$ \\ (36.1\%)\end{tabular} \\
EMEA (pt$\rightarrow$en) & \cellcolor[HTML]{F1FAEE} \begin{tabular}{@{}c@{}}-0.05 BLEU \\ (4.1\%)\end{tabular} & \cellcolor[HTML]{DEF2D9} \begin{tabular}{@{}c@{}}-0.20 BLEU$^{\ddagger}$ \\ (15.5\%)\end{tabular} & \cellcolor[HTML]{D3EECD} \begin{tabular}{@{}c@{}}+0.15 BLEU$^{\ddagger}$ \\ (20.0\%)\end{tabular} & \cellcolor[HTML]{F0F9EC} \begin{tabular}{@{}c@{}}-0.06 BLEU \\ (5.2\%)\end{tabular} & \cellcolor[HTML]{E3F4DE} \begin{tabular}{@{}c@{}}-0.15 BLEU$^{\ddagger}$ \\ (13.6\%)\end{tabular} & \cellcolor[HTML]{D5EFCF} \begin{tabular}{@{}c@{}}-0.13 BLEU$^{\ddagger}$ \\ (19.4\%)\end{tabular} & \cellcolor[HTML]{C9EAC2} \begin{tabular}{@{}c@{}}+0.11 BLEU$^{\ddagger}$ \\ (24.4\%)\end{tabular} & \cellcolor[HTML]{7AC77B} \begin{tabular}{@{}c@{}}-0.17 BLEU$^{\ddagger}$ \\ (48.2\%)\end{tabular} \\
JRC-Acquis (de$\rightarrow$en) & \cellcolor[HTML]{D9F0D3} \begin{tabular}{@{}c@{}}+0.50 BLEU$^{\ddagger}$ \\ (17.8\%)\end{tabular} & \cellcolor[HTML]{BBE4B4} \begin{tabular}{@{}c@{}}-0.12 BLEU \\ (29.1\%)\end{tabular} & \cellcolor[HTML]{CBEBC5} \begin{tabular}{@{}c@{}}+0.41 BLEU$^{\ddagger}$ \\ (23.3\%)\end{tabular} & \cellcolor[HTML]{BBE4B4} \begin{tabular}{@{}c@{}}+0.11 BLEU$^{\ddagger}$ \\ (29.0\%)\end{tabular} & \cellcolor[HTML]{E8F6E4} \begin{tabular}{@{}c@{}}+0.01 BLEU \\ (10.3\%)\end{tabular} & \cellcolor[HTML]{BEE5B8} \begin{tabular}{@{}c@{}}-0.12 BLEU \\ (28.0\%)\end{tabular} & \cellcolor[HTML]{EDF8EA} \begin{tabular}{@{}c@{}}-0.15 BLEU$^{\ddagger}$ \\ (6.7\%)\end{tabular} & \cellcolor[HTML]{DEF2D9} \begin{tabular}{@{}c@{}}-0.06 BLEU$^{\ddagger}$ \\ (15.3\%)\end{tabular} \\
JRC-Acquis (en$\rightarrow$de) & \cellcolor[HTML]{D3EECD} \begin{tabular}{@{}c@{}}+0.34 BLEU$^{\ddagger}$ \\ (20.2\%)\end{tabular} & \cellcolor[HTML]{C9EAC2} \begin{tabular}{@{}c@{}}+0.05 BLEU \\ (24.4\%)\end{tabular} & \cellcolor[HTML]{B1E0AB} \begin{tabular}{@{}c@{}}+0.57 BLEU$^{\ddagger}$ \\ (32.2\%)\end{tabular} & \cellcolor[HTML]{E3F4DE} \begin{tabular}{@{}c@{}}+0.05 BLEU \\ (13.5\%)\end{tabular} & \cellcolor[HTML]{D2EDCC} \begin{tabular}{@{}c@{}}-0.02 BLEU \\ (20.7\%)\end{tabular} & \cellcolor[HTML]{A7DBA0} \begin{tabular}{@{}c@{}}+0.32 BLEU$^{\ddagger}$ \\ (35.9\%)\end{tabular} & \cellcolor[HTML]{D5EFCF} \begin{tabular}{@{}c@{}}+0.09 BLEU$^{\ddagger}$ \\ (19.5\%)\end{tabular} & \cellcolor[HTML]{C0E6B9} \begin{tabular}{@{}c@{}}-0.10 BLEU$^{\ddagger}$ \\ (27.7\%)\end{tabular} \\
JRC-Acquis (en$\rightarrow$es) & \cellcolor[HTML]{A9DCA3} \begin{tabular}{@{}c@{}}+0.82 BLEU$^{\ddagger}$ \\ (35.0\%)\end{tabular} & \cellcolor[HTML]{C3E7BC} \begin{tabular}{@{}c@{}}+0.37 BLEU$^{\ddagger}$ \\ (26.4\%)\end{tabular} & \cellcolor[HTML]{E2F4DD} \begin{tabular}{@{}c@{}}+0.09 BLEU$^{\ddagger}$ \\ (13.7\%)\end{tabular} & \cellcolor[HTML]{D1EDCB} \begin{tabular}{@{}c@{}}+0.41 BLEU$^{\ddagger}$ \\ (20.9\%)\end{tabular} & \cellcolor[HTML]{ECF8E8} \begin{tabular}{@{}c@{}}-0.05 BLEU \\ (8.2\%)\end{tabular} & \cellcolor[HTML]{CEECC8} \begin{tabular}{@{}c@{}}-0.10 BLEU \\ (22.0\%)\end{tabular} & \cellcolor[HTML]{BCE4B5} \begin{tabular}{@{}c@{}}-0.03 BLEU \\ (28.8\%)\end{tabular} & \cellcolor[HTML]{ACDEA6} \begin{tabular}{@{}c@{}}+0.10 BLEU$^{\ddagger}$ \\ (33.9\%)\end{tabular} \\
JRC-Acquis (es$\rightarrow$en) & \cellcolor[HTML]{ECF8E8} \begin{tabular}{@{}c@{}}+0.35 BLEU$^{\ddagger}$ \\ (8.1\%)\end{tabular} & \cellcolor[HTML]{DCF2D7} \begin{tabular}{@{}c@{}}+0.03 BLEU \\ (16.2\%)\end{tabular} & \cellcolor[HTML]{BCE4B5} \begin{tabular}{@{}c@{}}+0.41 BLEU$^{\ddagger}$ \\ (28.7\%)\end{tabular} & \cellcolor[HTML]{FFF0E9} \begin{tabular}{@{}c@{}}-0.09 BLEU$^{\ddagger}$ \\ (-2.7\%)\end{tabular} & \cellcolor[HTML]{D3EECD} \begin{tabular}{@{}c@{}}-0.10 BLEU$^{\ddagger}$ \\ (20.0\%)\end{tabular} & \cellcolor[HTML]{BDE5B6} \begin{tabular}{@{}c@{}}+0.06 BLEU$^{\ddagger}$ \\ (28.5\%)\end{tabular} & \cellcolor[HTML]{DDF2D8} \begin{tabular}{@{}c@{}}+0.04 BLEU \\ (15.8\%)\end{tabular} & \cellcolor[HTML]{D6EFD0} \begin{tabular}{@{}c@{}}+0.01 BLEU \\ (18.8\%)\end{tabular} \\
JRC-Acquis (en$\rightarrow$pt) & \cellcolor[HTML]{D7EFD1} \begin{tabular}{@{}c@{}}-0.04 BLEU \\ (18.7\%)\end{tabular} & \cellcolor[HTML]{F1FAEE} \begin{tabular}{@{}c@{}}-0.18 BLEU \\ (4.4\%)\end{tabular} & \cellcolor[HTML]{E9F7E5} \begin{tabular}{@{}c@{}}-0.07 BLEU \\ (10.1\%)\end{tabular} & \cellcolor[HTML]{FFEDE5} \begin{tabular}{@{}c@{}}-0.24 BLEU$^{\ddagger}$ \\ (-4.9\%)\end{tabular} & \cellcolor[HTML]{EBF7E7} \begin{tabular}{@{}c@{}}-0.33 BLEU$^{\ddagger}$ \\ (8.5\%)\end{tabular} & \cellcolor[HTML]{E7F6E2} \begin{tabular}{@{}c@{}}-0.22 BLEU$^{\ddagger}$ \\ (11.6\%)\end{tabular} & \cellcolor[HTML]{ECF8E8} \begin{tabular}{@{}c@{}}-0.03 BLEU \\ (7.6\%)\end{tabular} & \cellcolor[HTML]{E3F4DE} \begin{tabular}{@{}c@{}}+0.03 BLEU \\ (13.3\%)\end{tabular} \\
JRC-Acquis (pt$\rightarrow$en) & \cellcolor[HTML]{FFECE4} \begin{tabular}{@{}c@{}}-0.02 BLEU \\ (-5.3\%)\end{tabular} & \cellcolor[HTML]{FFECE4} \begin{tabular}{@{}c@{}}-0.01 BLEU$^{\ddagger}$ \\ (-5.4\%)\end{tabular} & \cellcolor[HTML]{F4FBF2} \begin{tabular}{@{}c@{}}-0.08 BLEU$^{\ddagger}$ \\ (2.2\%)\end{tabular} & \cellcolor[HTML]{FFF2EC} \begin{tabular}{@{}c@{}}-0.22 BLEU$^{\ddagger}$ \\ (-1.8\%)\end{tabular} & \cellcolor[HTML]{FFF1EA} \begin{tabular}{@{}c@{}}-0.05 BLEU \\ (-2.6\%)\end{tabular} & \cellcolor[HTML]{E4F5DF} \begin{tabular}{@{}c@{}}-0.41 BLEU$^{\ddagger}$ \\ (12.9\%)\end{tabular} & \cellcolor[HTML]{ECF8E8} \begin{tabular}{@{}c@{}}-0.04 BLEU \\ (8.1\%)\end{tabular} & \cellcolor[HTML]{E0F3DB} \begin{tabular}{@{}c@{}}-0.20 BLEU$^{\ddagger}$ \\ (14.7\%)\end{tabular} \\
KDE (de$\rightarrow$en) & \cellcolor[HTML]{D7EFD1} \begin{tabular}{@{}c@{}}+0.17 BLEU$^{\ddagger}$ \\ (18.5\%)\end{tabular} & \cellcolor[HTML]{92D28F} \begin{tabular}{@{}c@{}}+0.11 BLEU$^{\ddagger}$ \\ (41.7\%)\end{tabular} & \cellcolor[HTML]{B8E3B2} \begin{tabular}{@{}c@{}}+0.63 BLEU$^{\ddagger}$ \\ (29.7\%)\end{tabular} & \cellcolor[HTML]{E8F6E4} \begin{tabular}{@{}c@{}}-0.12 BLEU$^{\ddagger}$ \\ (10.3\%)\end{tabular} & \cellcolor[HTML]{E3F4DE} \begin{tabular}{@{}c@{}}-0.16 BLEU$^{\ddagger}$ \\ (13.6\%)\end{tabular} & \cellcolor[HTML]{B5E1AE} \begin{tabular}{@{}c@{}}+0.56 BLEU$^{\ddagger}$ \\ (30.9\%)\end{tabular} & \cellcolor[HTML]{E0F3DB} \begin{tabular}{@{}c@{}}+0.06 BLEU$^{\ddagger}$ \\ (14.5\%)\end{tabular} & \cellcolor[HTML]{DAF0D4} \begin{tabular}{@{}c@{}}+0.07 BLEU$^{\ddagger}$ \\ (17.3\%)\end{tabular} \\
KDE (en$\rightarrow$de) & \cellcolor[HTML]{E1F3DC} \begin{tabular}{@{}c@{}}+0.15 BLEU$^{\ddagger}$ \\ (14.2\%)\end{tabular} & \cellcolor[HTML]{FFF4EE} \begin{tabular}{@{}c@{}}-0.09 BLEU$^{\ddagger}$ \\ (-0.9\%)\end{tabular} & \cellcolor[HTML]{DCF2D7} \begin{tabular}{@{}c@{}}-0.01 BLEU \\ (16.2\%)\end{tabular} & \cellcolor[HTML]{EFF9EB} \begin{tabular}{@{}c@{}}-0.02 BLEU \\ (6.2\%)\end{tabular} & \cellcolor[HTML]{EFF9EB} \begin{tabular}{@{}c@{}}-0.14 BLEU \\ (5.9\%)\end{tabular} & \cellcolor[HTML]{EEF8EA} \begin{tabular}{@{}c@{}}-0.04 BLEU \\ (6.6\%)\end{tabular} & \cellcolor[HTML]{E6F5E1} \begin{tabular}{@{}c@{}}-0.09 BLEU$^{\ddagger}$ \\ (11.9\%)\end{tabular} & \cellcolor[HTML]{E3F4DE} \begin{tabular}{@{}c@{}}-0.05 BLEU$^{\ddagger}$ \\ (13.6\%)\end{tabular} \\
KDE (en$\rightarrow$es) & \cellcolor[HTML]{D8F0D2} \begin{tabular}{@{}c@{}}+0.28 BLEU$^{\ddagger}$ \\ (18.1\%)\end{tabular} & \cellcolor[HTML]{E5F5E1} \begin{tabular}{@{}c@{}}+0.03 BLEU \\ (12.3\%)\end{tabular} & \cellcolor[HTML]{E0F3DB} \begin{tabular}{@{}c@{}}+0.01 BLEU \\ (14.8\%)\end{tabular} & \cellcolor[HTML]{EDF8EA} \begin{tabular}{@{}c@{}}-0.23 BLEU$^{\ddagger}$ \\ (6.8\%)\end{tabular} & \cellcolor[HTML]{CDECC7} \begin{tabular}{@{}c@{}}-0.24 BLEU$^{\ddagger}$ \\ (22.4\%)\end{tabular} & \cellcolor[HTML]{E2F4DD} \begin{tabular}{@{}c@{}}-0.01 BLEU \\ (14.1\%)\end{tabular} & \cellcolor[HTML]{DBF1D6} \begin{tabular}{@{}c@{}}-0.16 BLEU$^{\ddagger}$ \\ (16.5\%)\end{tabular} & \cellcolor[HTML]{E5F5E1} \begin{tabular}{@{}c@{}}-0.04 BLEU \\ (12.4\%)\end{tabular} \\
KDE (es$\rightarrow$en) & \cellcolor[HTML]{CCEBC6} \begin{tabular}{@{}c@{}}+0.40 BLEU$^{\ddagger}$ \\ (23.0\%)\end{tabular} & \cellcolor[HTML]{D5EFCF} \begin{tabular}{@{}c@{}}-0.09 BLEU \\ (19.2\%)\end{tabular} & \cellcolor[HTML]{CFECC9} \begin{tabular}{@{}c@{}}+0.19 BLEU$^{\ddagger}$ \\ (21.6\%)\end{tabular} & \cellcolor[HTML]{ECF8E8} \begin{tabular}{@{}c@{}}-0.18 BLEU$^{\ddagger}$ \\ (7.7\%)\end{tabular} & \cellcolor[HTML]{E0F3DB} \begin{tabular}{@{}c@{}}+0.06 BLEU \\ (14.5\%)\end{tabular} & \cellcolor[HTML]{BCE4B5} \begin{tabular}{@{}c@{}}-0.16 BLEU \\ (28.5\%)\end{tabular} & \cellcolor[HTML]{A4DA9E} \begin{tabular}{@{}c@{}}-0.33 BLEU$^{\ddagger}$ \\ (36.4\%)\end{tabular} & \cellcolor[HTML]{D8F0D2} \begin{tabular}{@{}c@{}}+0.01 BLEU \\ (18.1\%)\end{tabular} \\
KDE (en$\rightarrow$pt) & \cellcolor[HTML]{C7E9C0} \begin{tabular}{@{}c@{}}+0.25 BLEU$^{\ddagger}$ \\ (25.3\%)\end{tabular} & \cellcolor[HTML]{C7E9C0} \begin{tabular}{@{}c@{}}+0.16 BLEU$^{\ddagger}$ \\ (25.3\%)\end{tabular} & \cellcolor[HTML]{F2FAF0} \begin{tabular}{@{}c@{}}+0.10 BLEU$^{\ddagger}$ \\ (3.1\%)\end{tabular} & \cellcolor[HTML]{F7FCF5} \begin{tabular}{@{}c@{}}-0.09 BLEU \\ (0.3\%)\end{tabular} & \cellcolor[HTML]{F6FCF4} \begin{tabular}{@{}c@{}}-0.15 BLEU$^{\ddagger}$ \\ (1.1\%)\end{tabular} & \cellcolor[HTML]{DDF2D8} \begin{tabular}{@{}c@{}}-0.02 BLEU \\ (15.9\%)\end{tabular} & \cellcolor[HTML]{E6F5E1} \begin{tabular}{@{}c@{}}-0.07 BLEU \\ (11.9\%)\end{tabular} & \cellcolor[HTML]{BBE4B4} \begin{tabular}{@{}c@{}}+0.00 BLEU \\ (29.2\%)\end{tabular} \\
KDE (pt$\rightarrow$en) & \cellcolor[HTML]{9CD797} \begin{tabular}{@{}c@{}}+0.02 BLEU \\ (38.9\%)\end{tabular} & \cellcolor[HTML]{AEDEA7} \begin{tabular}{@{}c@{}}-0.31 BLEU$^{\ddagger}$ \\ (33.6\%)\end{tabular} & \cellcolor[HTML]{ABDDA5} \begin{tabular}{@{}c@{}}+0.08 BLEU \\ (34.0\%)\end{tabular} & \cellcolor[HTML]{F1FAEE} \begin{tabular}{@{}c@{}}-0.05 BLEU \\ (4.0\%)\end{tabular} & \cellcolor[HTML]{D8F0D2} \begin{tabular}{@{}c@{}}+0.08 BLEU \\ (18.1\%)\end{tabular} & \cellcolor[HTML]{AFDFA8} \begin{tabular}{@{}c@{}}-0.14 BLEU$^{\ddagger}$ \\ (32.8\%)\end{tabular} & \cellcolor[HTML]{D6EFD0} \begin{tabular}{@{}c@{}}-0.01 BLEU \\ (19.0\%)\end{tabular} & \cellcolor[HTML]{CAEAC3} \begin{tabular}{@{}c@{}}-0.03 BLEU \\ (23.8\%)\end{tabular} \\
NLU\_AgNews & \cellcolor[HTML]{289049}\color{white} \begin{tabular}{@{}c@{}}+0.01 Acc \\ (72.9\%)\end{tabular} & \cellcolor[HTML]{38A156}\color{white} \begin{tabular}{@{}c@{}}-0.39 Acc$^{\ddagger}$ \\ (66.2\%)\end{tabular} & \cellcolor[HTML]{79C67A} \begin{tabular}{@{}c@{}}-0.14 Acc \\ (48.7\%)\end{tabular} & \cellcolor[HTML]{137D39}\color{white} \begin{tabular}{@{}c@{}}-0.00 Acc \\ (80.7\%)\end{tabular} & \cellcolor[HTML]{8BCF89} \begin{tabular}{@{}c@{}}-0.11 Acc$^{\ddagger}$ \\ (43.7\%)\end{tabular} & \cellcolor[HTML]{52B365} \begin{tabular}{@{}c@{}}-0.26 Acc$^{\ddagger}$ \\ (58.6\%)\end{tabular} & \cellcolor[HTML]{5DB96B} \begin{tabular}{@{}c@{}}-0.36 Acc$^{\ddagger}$ \\ (55.7\%)\end{tabular} & \cellcolor[HTML]{97D492} \begin{tabular}{@{}c@{}}-0.30 Acc$^{\ddagger}$ \\ (40.5\%)\end{tabular} \\
NLU\_CR & \cellcolor[HTML]{38A156}\color{white} \begin{tabular}{@{}c@{}}+0.11 Acc \\ (66.2\%)\end{tabular} & \cellcolor[HTML]{2A924A}\color{white} \begin{tabular}{@{}c@{}}+1.20 Acc$^{\ddagger}$ \\ (71.9\%)\end{tabular} & \cellcolor[HTML]{006027}\color{white} \begin{tabular}{@{}c@{}}+0.43 Acc$^{\ddagger}$ \\ (91.2\%)\end{tabular} & \cellcolor[HTML]{1A843F}\color{white} \begin{tabular}{@{}c@{}}-0.32 Acc \\ (78.0\%)\end{tabular} & \cellcolor[HTML]{006529}\color{white} \begin{tabular}{@{}c@{}}+0.13 Acc \\ (89.6\%)\end{tabular} & \cellcolor[HTML]{17813D}\color{white} \begin{tabular}{@{}c@{}}-0.61 Acc$^{\ddagger}$ \\ (79.0\%)\end{tabular} & \cellcolor[HTML]{0C7735}\color{white} \begin{tabular}{@{}c@{}}+0.21 Acc \\ (83.1\%)\end{tabular} & \cellcolor[HTML]{097532}\color{white} \begin{tabular}{@{}c@{}}+0.05 Acc \\ (84.0\%)\end{tabular} \\
NLU\_MNLI & \cellcolor[HTML]{A9DCA3} \begin{tabular}{@{}c@{}}+0.29 Acc \\ (35.0\%)\end{tabular} & \cellcolor[HTML]{88CE87} \begin{tabular}{@{}c@{}}+0.10 Acc \\ (44.3\%)\end{tabular} & \cellcolor[HTML]{62BB6D} \begin{tabular}{@{}c@{}}-0.61 Acc$^{\ddagger}$ \\ (54.4\%)\end{tabular} & \cellcolor[HTML]{90D18D} \begin{tabular}{@{}c@{}}-0.23 Acc$^{\ddagger}$ \\ (42.3\%)\end{tabular} & \cellcolor[HTML]{83CB82} \begin{tabular}{@{}c@{}}-0.50 Acc$^{\ddagger}$ \\ (46.0\%)\end{tabular} & \cellcolor[HTML]{43AC5E}\color{white} \begin{tabular}{@{}c@{}}-0.23 Acc \\ (62.0\%)\end{tabular} & \cellcolor[HTML]{B1E0AB} \begin{tabular}{@{}c@{}}-0.19 Acc$^{\ddagger}$ \\ (32.3\%)\end{tabular} & \cellcolor[HTML]{90D18D} \begin{tabular}{@{}c@{}}-1.06 Acc$^{\ddagger}$ \\ (42.5\%)\end{tabular} \\
NLU\_QNLI & \cellcolor[HTML]{A8DCA2} \begin{tabular}{@{}c@{}}+0.93 Acc$^{\ddagger}$ \\ (35.3\%)\end{tabular} & \cellcolor[HTML]{D6EFD0} \begin{tabular}{@{}c@{}}-0.80 Acc$^{\ddagger}$ \\ (18.8\%)\end{tabular} & \cellcolor[HTML]{349D53}\color{white} \begin{tabular}{@{}c@{}}-0.91 Acc$^{\ddagger}$ \\ (67.6\%)\end{tabular} & \cellcolor[HTML]{248C46}\color{white} \begin{tabular}{@{}c@{}}+1.35 Acc$^{\ddagger}$ \\ (74.3\%)\end{tabular} & \cellcolor[HTML]{C0E6B9} \begin{tabular}{@{}c@{}}-0.05 Acc \\ (27.5\%)\end{tabular} & \cellcolor[HTML]{B7E2B1} \begin{tabular}{@{}c@{}}-0.28 Acc$^{\ddagger}$ \\ (30.4\%)\end{tabular} & \cellcolor[HTML]{349D53}\color{white} \begin{tabular}{@{}c@{}}+0.01 Acc \\ (67.7\%)\end{tabular} & \cellcolor[HTML]{70C274} \begin{tabular}{@{}c@{}}+0.84 Acc$^{\ddagger}$ \\ (51.0\%)\end{tabular} \\
NLU\_SST-2 & \cellcolor[HTML]{AADDA4} \begin{tabular}{@{}c@{}}-0.36 Acc$^{\ddagger}$ \\ (34.6\%)\end{tabular} & \cellcolor[HTML]{99D595} \begin{tabular}{@{}c@{}}-0.48 Acc$^{\ddagger}$ \\ (39.8\%)\end{tabular} & \cellcolor[HTML]{005C25}\color{white} \begin{tabular}{@{}c@{}}+0.09 Acc \\ (92.5\%)\end{tabular} & \cellcolor[HTML]{3DA65A}\color{white} \begin{tabular}{@{}c@{}}+0.05 Acc \\ (64.3\%)\end{tabular} & \cellcolor[HTML]{4EB264} \begin{tabular}{@{}c@{}}-0.30 Acc$^{\ddagger}$ \\ (59.2\%)\end{tabular} & \cellcolor[HTML]{258D47}\color{white} \begin{tabular}{@{}c@{}}-0.22 Acc$^{\ddagger}$ \\ (74.0\%)\end{tabular} & \cellcolor[HTML]{3CA559}\color{white} \begin{tabular}{@{}c@{}}+0.34 Acc$^{\ddagger}$ \\ (64.7\%)\end{tabular} & \cellcolor[HTML]{8BCF89} \begin{tabular}{@{}c@{}}-0.28 Acc$^{\ddagger}$ \\ (43.7\%)\end{tabular} \\
NLU\_SST-5 & \cellcolor[HTML]{2E964D}\color{white} \begin{tabular}{@{}c@{}}+1.95 Acc$^{\ddagger}$ \\ (70.4\%)\end{tabular} & \cellcolor[HTML]{6BC072} \begin{tabular}{@{}c@{}}-0.71 Acc$^{\ddagger}$ \\ (52.2\%)\end{tabular} & \cellcolor[HTML]{53B466} \begin{tabular}{@{}c@{}}+0.87 Acc$^{\ddagger}$ \\ (57.9\%)\end{tabular} & \cellcolor[HTML]{4EB264} \begin{tabular}{@{}c@{}}+0.65 Acc$^{\ddagger}$ \\ (59.0\%)\end{tabular} & \cellcolor[HTML]{56B567} \begin{tabular}{@{}c@{}}+1.68 Acc$^{\ddagger}$ \\ (57.3\%)\end{tabular} & \cellcolor[HTML]{2E964D}\color{white} \begin{tabular}{@{}c@{}}+1.51 Acc$^{\ddagger}$ \\ (70.4\%)\end{tabular} & \cellcolor[HTML]{46AE60}\color{white} \begin{tabular}{@{}c@{}}+0.47 Acc$^{\ddagger}$ \\ (61.1\%)\end{tabular} & \cellcolor[HTML]{60BA6C} \begin{tabular}{@{}c@{}}+0.87 Acc$^{\ddagger}$ \\ (55.0\%)\end{tabular} \\
NLU\_Subj & \cellcolor[HTML]{359E53}\color{white} \begin{tabular}{@{}c@{}}+1.25 Acc$^{\ddagger}$ \\ (67.3\%)\end{tabular} & \cellcolor[HTML]{C0E6B9} \begin{tabular}{@{}c@{}}-0.74 Acc$^{\ddagger}$ \\ (27.4\%)\end{tabular} & \cellcolor[HTML]{E9F7E5} \begin{tabular}{@{}c@{}}-0.43 Acc$^{\ddagger}$ \\ (10.2\%)\end{tabular} & \cellcolor[HTML]{1E8741}\color{white} \begin{tabular}{@{}c@{}}+4.02 Acc$^{\ddagger}$ \\ (76.6\%)\end{tabular} & \cellcolor[HTML]{D8F0D2} \begin{tabular}{@{}c@{}}-0.40 Acc \\ (18.3\%)\end{tabular} & \cellcolor[HTML]{E7F6E2} \begin{tabular}{@{}c@{}}-0.21 Acc$^{\ddagger}$ \\ (11.5\%)\end{tabular} & \cellcolor[HTML]{E7F6E3} \begin{tabular}{@{}c@{}}-0.12 Acc \\ (10.9\%)\end{tabular} & \cellcolor[HTML]{D4EECE} \begin{tabular}{@{}c@{}}-0.28 Acc$^{\ddagger}$ \\ (19.7\%)\end{tabular} \\
\hline
\end{tabular}%
   }
\caption{Performance gains (BLEU/Accuracy) and token savings (\%) evaluated using the Gating (dev-set calibrated) configuration. Cell colors represent the percentage of tokens saved. $^{\ddagger}$ denotes statistical significance vs.\ full reranking ($p<0.05$, two-sided).}
\label{tab:heatmap_results_dev_calibrated}
\end{table*}

\begin{table*}[t]
\centering
\resizebox{\textwidth}{!}{%
\begin{tabular}{l c c c c c c c c}
\hline
Dataset & Qwen 3B & Llama 3B & Command & Mistral & Aya & Llama 8B & Qwen 32B & Llama 70B \\
\hline
EMEA (de$\rightarrow$en) & \cellcolor[HTML]{DEF2D9} \begin{tabular}{@{}c@{}}+0.65 BLEU$^{\ddagger}$ \\ (15.3\%)\end{tabular} & \cellcolor[HTML]{CAEAC3} \begin{tabular}{@{}c@{}}+0.54 BLEU$^{\ddagger}$ \\ (23.9\%)\end{tabular} & \cellcolor[HTML]{E1F3DC} \begin{tabular}{@{}c@{}}+0.96 BLEU$^{\ddagger}$ \\ (14.1\%)\end{tabular} & \cellcolor[HTML]{EAF7E6} \begin{tabular}{@{}c@{}}+0.48 BLEU$^{\ddagger}$ \\ (9.1\%)\end{tabular} & \cellcolor[HTML]{CEECC8} \begin{tabular}{@{}c@{}}+0.26 BLEU$^{\ddagger}$ \\ (22.1\%)\end{tabular} & \cellcolor[HTML]{A8DCA2} \begin{tabular}{@{}c@{}}+0.51 BLEU$^{\ddagger}$ \\ (35.2\%)\end{tabular} & \cellcolor[HTML]{C3E7BC} \begin{tabular}{@{}c@{}}+0.49 BLEU$^{\ddagger}$ \\ (26.3\%)\end{tabular} & \cellcolor[HTML]{E4F5DF} \begin{tabular}{@{}c@{}}+0.03 BLEU \\ (12.9\%)\end{tabular} \\
EMEA (en$\rightarrow$de) & \cellcolor[HTML]{C1E6BA} \begin{tabular}{@{}c@{}}+0.69 BLEU$^{\ddagger}$ \\ (27.3\%)\end{tabular} & \cellcolor[HTML]{CDECC7} \begin{tabular}{@{}c@{}}+0.64 BLEU$^{\ddagger}$ \\ (22.3\%)\end{tabular} & \cellcolor[HTML]{EFF9EB} \begin{tabular}{@{}c@{}}+0.22 BLEU$^{\ddagger}$ \\ (6.2\%)\end{tabular} & \cellcolor[HTML]{58B668} \begin{tabular}{@{}c@{}}+0.48 BLEU$^{\ddagger}$ \\ (56.7\%)\end{tabular} & \cellcolor[HTML]{A8DCA2} \begin{tabular}{@{}c@{}}+0.71 BLEU$^{\ddagger}$ \\ (35.4\%)\end{tabular} & \cellcolor[HTML]{CAEAC3} \begin{tabular}{@{}c@{}}+0.94 BLEU$^{\ddagger}$ \\ (24.1\%)\end{tabular} & \cellcolor[HTML]{D9F0D3} \begin{tabular}{@{}c@{}}+0.16 BLEU$^{\ddagger}$ \\ (17.6\%)\end{tabular} & \cellcolor[HTML]{EFF9EC} \begin{tabular}{@{}c@{}}+0.07 BLEU$^{\ddagger}$ \\ (5.6\%)\end{tabular} \\
EMEA (en$\rightarrow$es) & \cellcolor[HTML]{C6E8BF} \begin{tabular}{@{}c@{}}+0.61 BLEU$^{\ddagger}$ \\ (25.7\%)\end{tabular} & \cellcolor[HTML]{86CC85} \begin{tabular}{@{}c@{}}+0.40 BLEU$^{\ddagger}$ \\ (45.2\%)\end{tabular} & \cellcolor[HTML]{DAF0D4} \begin{tabular}{@{}c@{}}+0.55 BLEU$^{\ddagger}$ \\ (17.2\%)\end{tabular} & \cellcolor[HTML]{D1EDCB} \begin{tabular}{@{}c@{}}+0.33 BLEU$^{\ddagger}$ \\ (21.0\%)\end{tabular} & \cellcolor[HTML]{79C67A} \begin{tabular}{@{}c@{}}+0.64 BLEU$^{\ddagger}$ \\ (48.6\%)\end{tabular} & \cellcolor[HTML]{9FD899} \begin{tabular}{@{}c@{}}+1.01 BLEU$^{\ddagger}$ \\ (38.0\%)\end{tabular} & \cellcolor[HTML]{CCEBC6} \begin{tabular}{@{}c@{}}+0.19 BLEU$^{\ddagger}$ \\ (22.9\%)\end{tabular} & \cellcolor[HTML]{D9F0D3} \begin{tabular}{@{}c@{}}+0.22 BLEU$^{\ddagger}$ \\ (17.6\%)\end{tabular} \\
EMEA (es$\rightarrow$en) & \cellcolor[HTML]{AFDFA8} \begin{tabular}{@{}c@{}}+0.69 BLEU$^{\ddagger}$ \\ (33.0\%)\end{tabular} & \cellcolor[HTML]{B2E0AC} \begin{tabular}{@{}c@{}}+0.69 BLEU$^{\ddagger}$ \\ (31.8\%)\end{tabular} & \cellcolor[HTML]{9FD899} \begin{tabular}{@{}c@{}}+1.42 BLEU$^{\ddagger}$ \\ (38.0\%)\end{tabular} & \cellcolor[HTML]{F1FAEE} \begin{tabular}{@{}c@{}}+0.04 BLEU$^{\ddagger}$ \\ (4.1\%)\end{tabular} & \cellcolor[HTML]{F1FAEE} \begin{tabular}{@{}c@{}}+0.52 BLEU$^{\ddagger}$ \\ (4.2\%)\end{tabular} & \cellcolor[HTML]{AFDFA8} \begin{tabular}{@{}c@{}}+0.56 BLEU$^{\ddagger}$ \\ (33.0\%)\end{tabular} & \cellcolor[HTML]{C2E7BB} \begin{tabular}{@{}c@{}}+0.20 BLEU$^{\ddagger}$ \\ (26.8\%)\end{tabular} & \cellcolor[HTML]{D1EDCB} \begin{tabular}{@{}c@{}}-0.04 BLEU$^{\ddagger}$ \\ (20.8\%)\end{tabular} \\
EMEA (en$\rightarrow$pt) & \cellcolor[HTML]{CAEAC3} \begin{tabular}{@{}c@{}}+0.94 BLEU$^{\ddagger}$ \\ (24.0\%)\end{tabular} & \cellcolor[HTML]{D6EFD0} \begin{tabular}{@{}c@{}}+0.25 BLEU$^{\ddagger}$ \\ (19.0\%)\end{tabular} & \cellcolor[HTML]{C3E7BC} \begin{tabular}{@{}c@{}}+2.53 BLEU$^{\ddagger}$ \\ (26.3\%)\end{tabular} & \cellcolor[HTML]{309950}\color{white} \begin{tabular}{@{}c@{}}+1.85 BLEU$^{\ddagger}$ \\ (69.1\%)\end{tabular} & \cellcolor[HTML]{2B934B}\color{white} \begin{tabular}{@{}c@{}}+0.64 BLEU$^{\ddagger}$ \\ (71.5\%)\end{tabular} & \cellcolor[HTML]{6DC072} \begin{tabular}{@{}c@{}}+1.74 BLEU$^{\ddagger}$ \\ (51.6\%)\end{tabular} & \cellcolor[HTML]{C1E6BA} \begin{tabular}{@{}c@{}}+0.66 BLEU$^{\ddagger}$ \\ (27.2\%)\end{tabular} & \cellcolor[HTML]{B0DFAA} \begin{tabular}{@{}c@{}}+1.44 BLEU$^{\ddagger}$ \\ (32.4\%)\end{tabular} \\
EMEA (pt$\rightarrow$en) & \cellcolor[HTML]{EFF9EC} \begin{tabular}{@{}c@{}}+0.08 BLEU$^{\ddagger}$ \\ (5.8\%)\end{tabular} & \cellcolor[HTML]{E7F6E2} \begin{tabular}{@{}c@{}}+0.14 BLEU$^{\ddagger}$ \\ (11.3\%)\end{tabular} & \cellcolor[HTML]{D8F0D2} \begin{tabular}{@{}c@{}}+0.52 BLEU$^{\ddagger}$ \\ (18.3\%)\end{tabular} & \cellcolor[HTML]{FFF4EF} \begin{tabular}{@{}c@{}}+0.17 BLEU$^{\ddagger}$ \\ (-0.5\%)\end{tabular} & \cellcolor[HTML]{FFF4EE} \begin{tabular}{@{}c@{}}+0.16 BLEU$^{\ddagger}$ \\ (-1.1\%)\end{tabular} & \cellcolor[HTML]{DAF0D4} \begin{tabular}{@{}c@{}}+0.15 BLEU$^{\ddagger}$ \\ (17.5\%)\end{tabular} & \cellcolor[HTML]{CDECC7} \begin{tabular}{@{}c@{}}+0.27 BLEU$^{\ddagger}$ \\ (22.6\%)\end{tabular} & \cellcolor[HTML]{6EC173} \begin{tabular}{@{}c@{}}+0.12 BLEU$^{\ddagger}$ \\ (51.3\%)\end{tabular} \\
JRC-Acquis (de$\rightarrow$en) & \cellcolor[HTML]{E6F5E1} \begin{tabular}{@{}c@{}}+0.60 BLEU$^{\ddagger}$ \\ (11.8\%)\end{tabular} & \cellcolor[HTML]{CBEBC5} \begin{tabular}{@{}c@{}}+0.23 BLEU$^{\ddagger}$ \\ (23.2\%)\end{tabular} & \cellcolor[HTML]{D5EFCF} \begin{tabular}{@{}c@{}}+0.62 BLEU$^{\ddagger}$ \\ (19.2\%)\end{tabular} & \cellcolor[HTML]{A5DB9F} \begin{tabular}{@{}c@{}}+0.32 BLEU$^{\ddagger}$ \\ (36.1\%)\end{tabular} & \cellcolor[HTML]{F2FAF0} \begin{tabular}{@{}c@{}}+0.20 BLEU$^{\ddagger}$ \\ (3.1\%)\end{tabular} & \cellcolor[HTML]{D5EFCF} \begin{tabular}{@{}c@{}}+0.14 BLEU$^{\ddagger}$ \\ (19.5\%)\end{tabular} & \cellcolor[HTML]{F7FCF5} \begin{tabular}{@{}c@{}}-0.02 BLEU$^{\ddagger}$ \\ (0.1\%)\end{tabular} & \cellcolor[HTML]{E8F6E3} \begin{tabular}{@{}c@{}}+0.03 BLEU$^{\ddagger}$ \\ (10.8\%)\end{tabular} \\
JRC-Acquis (en$\rightarrow$de) & \cellcolor[HTML]{D0EDCA} \begin{tabular}{@{}c@{}}+0.54 BLEU$^{\ddagger}$ \\ (21.2\%)\end{tabular} & \cellcolor[HTML]{CDECC7} \begin{tabular}{@{}c@{}}+0.29 BLEU$^{\ddagger}$ \\ (22.5\%)\end{tabular} & \cellcolor[HTML]{A9DCA3} \begin{tabular}{@{}c@{}}+0.74 BLEU$^{\ddagger}$ \\ (34.8\%)\end{tabular} & \cellcolor[HTML]{F0F9ED} \begin{tabular}{@{}c@{}}+0.32 BLEU$^{\ddagger}$ \\ (4.7\%)\end{tabular} & \cellcolor[HTML]{D4EECE} \begin{tabular}{@{}c@{}}+0.21 BLEU$^{\ddagger}$ \\ (19.6\%)\end{tabular} & \cellcolor[HTML]{B5E1AE} \begin{tabular}{@{}c@{}}+0.59 BLEU$^{\ddagger}$ \\ (30.9\%)\end{tabular} & \cellcolor[HTML]{CEECC8} \begin{tabular}{@{}c@{}}+0.27 BLEU$^{\ddagger}$ \\ (21.9\%)\end{tabular} & \cellcolor[HTML]{E7F6E2} \begin{tabular}{@{}c@{}}+0.15 BLEU$^{\ddagger}$ \\ (11.4\%)\end{tabular} \\
JRC-Acquis (en$\rightarrow$es) & \cellcolor[HTML]{ACDEA6} \begin{tabular}{@{}c@{}}+1.05 BLEU$^{\ddagger}$ \\ (33.9\%)\end{tabular} & \cellcolor[HTML]{D8F0D2} \begin{tabular}{@{}c@{}}+0.58 BLEU$^{\ddagger}$ \\ (18.0\%)\end{tabular} & \cellcolor[HTML]{EBF7E7} \begin{tabular}{@{}c@{}}+0.31 BLEU$^{\ddagger}$ \\ (8.9\%)\end{tabular} & \cellcolor[HTML]{D8F0D2} \begin{tabular}{@{}c@{}}+0.59 BLEU$^{\ddagger}$ \\ (18.2\%)\end{tabular} & \cellcolor[HTML]{F4FBF2} \begin{tabular}{@{}c@{}}+0.17 BLEU$^{\ddagger}$ \\ (2.3\%)\end{tabular} & \cellcolor[HTML]{E3F4DE} \begin{tabular}{@{}c@{}}+0.28 BLEU$^{\ddagger}$ \\ (13.3\%)\end{tabular} & \cellcolor[HTML]{D1EDCB} \begin{tabular}{@{}c@{}}+0.27 BLEU$^{\ddagger}$ \\ (20.8\%)\end{tabular} & \cellcolor[HTML]{CFECC9} \begin{tabular}{@{}c@{}}+0.32 BLEU$^{\ddagger}$ \\ (21.7\%)\end{tabular} \\
JRC-Acquis (es$\rightarrow$en) & \cellcolor[HTML]{ECF8E8} \begin{tabular}{@{}c@{}}+0.54 BLEU$^{\ddagger}$ \\ (8.0\%)\end{tabular} & \cellcolor[HTML]{DBF1D6} \begin{tabular}{@{}c@{}}+0.28 BLEU$^{\ddagger}$ \\ (16.6\%)\end{tabular} & \cellcolor[HTML]{C1E6BA} \begin{tabular}{@{}c@{}}+0.55 BLEU$^{\ddagger}$ \\ (27.2\%)\end{tabular} & \cellcolor[HTML]{FFECE4} \begin{tabular}{@{}c@{}}+0.00 BLEU \\ (-5.4\%)\end{tabular} & \cellcolor[HTML]{EAF7E6} \begin{tabular}{@{}c@{}}+0.07 BLEU$^{\ddagger}$ \\ (9.3\%)\end{tabular} & \cellcolor[HTML]{A4DA9E} \begin{tabular}{@{}c@{}}+0.27 BLEU$^{\ddagger}$ \\ (36.4\%)\end{tabular} & \cellcolor[HTML]{DAF0D4} \begin{tabular}{@{}c@{}}+0.19 BLEU$^{\ddagger}$ \\ (17.4\%)\end{tabular} & \cellcolor[HTML]{E8F6E4} \begin{tabular}{@{}c@{}}+0.10 BLEU$^{\ddagger}$ \\ (10.4\%)\end{tabular} \\
JRC-Acquis (en$\rightarrow$pt) & \cellcolor[HTML]{DBF1D5} \begin{tabular}{@{}c@{}}+0.21 BLEU$^{\ddagger}$ \\ (17.1\%)\end{tabular} & \cellcolor[HTML]{FFF2EB} \begin{tabular}{@{}c@{}}+0.15 BLEU$^{\ddagger}$ \\ (-2.0\%)\end{tabular} & \cellcolor[HTML]{E0F3DB} \begin{tabular}{@{}c@{}}+0.15 BLEU$^{\ddagger}$ \\ (14.8\%)\end{tabular} & \cellcolor[HTML]{FEE8DD} \begin{tabular}{@{}c@{}}-0.15 BLEU$^{\ddagger}$ \\ (-8.1\%)\end{tabular} & \cellcolor[HTML]{FFEDE5} \begin{tabular}{@{}c@{}}-0.03 BLEU$^{\ddagger}$ \\ (-4.7\%)\end{tabular} & \cellcolor[HTML]{F5FBF2} \begin{tabular}{@{}c@{}}+0.05 BLEU$^{\ddagger}$ \\ (1.6\%)\end{tabular} & \cellcolor[HTML]{EEF8EA} \begin{tabular}{@{}c@{}}+0.15 BLEU$^{\ddagger}$ \\ (6.5\%)\end{tabular} & \cellcolor[HTML]{ECF8E8} \begin{tabular}{@{}c@{}}+0.18 BLEU$^{\ddagger}$ \\ (7.6\%)\end{tabular} \\
JRC-Acquis (pt$\rightarrow$en) & \cellcolor[HTML]{FFECE3} \begin{tabular}{@{}c@{}}-0.00 BLEU \\ (-5.5\%)\end{tabular} & \cellcolor[HTML]{FFECE3} \begin{tabular}{@{}c@{}}+0.00 BLEU \\ (-5.7\%)\end{tabular} & \cellcolor[HTML]{F4FBF1} \begin{tabular}{@{}c@{}}+0.03 BLEU$^{\ddagger}$ \\ (2.5\%)\end{tabular} & \cellcolor[HTML]{FFEDE5} \begin{tabular}{@{}c@{}}+0.00 BLEU \\ (-4.9\%)\end{tabular} & \cellcolor[HTML]{FFECE3} \begin{tabular}{@{}c@{}}+0.00 BLEU \\ (-5.7\%)\end{tabular} & \cellcolor[HTML]{FFECE3} \begin{tabular}{@{}c@{}}+0.00 BLEU \\ (-5.6\%)\end{tabular} & \cellcolor[HTML]{F2FAF0} \begin{tabular}{@{}c@{}}+0.10 BLEU$^{\ddagger}$ \\ (3.4\%)\end{tabular} & \cellcolor[HTML]{F1FAEE} \begin{tabular}{@{}c@{}}-0.03 BLEU$^{\ddagger}$ \\ (4.4\%)\end{tabular} \\
KDE (de$\rightarrow$en) & \cellcolor[HTML]{C3E7BC} \begin{tabular}{@{}c@{}}+0.39 BLEU$^{\ddagger}$ \\ (26.6\%)\end{tabular} & \cellcolor[HTML]{60BA6C} \begin{tabular}{@{}c@{}}+0.46 BLEU$^{\ddagger}$ \\ (54.7\%)\end{tabular} & \cellcolor[HTML]{C7E9C0} \begin{tabular}{@{}c@{}}+0.89 BLEU$^{\ddagger}$ \\ (25.3\%)\end{tabular} & \cellcolor[HTML]{FFF4EE} \begin{tabular}{@{}c@{}}+0.16 BLEU$^{\ddagger}$ \\ (-1.0\%)\end{tabular} & \cellcolor[HTML]{EDF8EA} \begin{tabular}{@{}c@{}}+0.07 BLEU$^{\ddagger}$ \\ (6.7\%)\end{tabular} & \cellcolor[HTML]{CCEBC6} \begin{tabular}{@{}c@{}}+0.89 BLEU$^{\ddagger}$ \\ (22.7\%)\end{tabular} & \cellcolor[HTML]{DFF3DA} \begin{tabular}{@{}c@{}}+0.17 BLEU$^{\ddagger}$ \\ (14.9\%)\end{tabular} & \cellcolor[HTML]{D8F0D2} \begin{tabular}{@{}c@{}}+0.24 BLEU$^{\ddagger}$ \\ (18.3\%)\end{tabular} \\
KDE (en$\rightarrow$de) & \cellcolor[HTML]{EBF7E7} \begin{tabular}{@{}c@{}}+0.46 BLEU$^{\ddagger}$ \\ (8.9\%)\end{tabular} & \cellcolor[HTML]{F5FBF2} \begin{tabular}{@{}c@{}}+0.06 BLEU$^{\ddagger}$ \\ (1.7\%)\end{tabular} & \cellcolor[HTML]{C6E8BF} \begin{tabular}{@{}c@{}}+0.30 BLEU$^{\ddagger}$ \\ (25.6\%)\end{tabular} & \cellcolor[HTML]{FFEFE8} \begin{tabular}{@{}c@{}}+0.27 BLEU$^{\ddagger}$ \\ (-3.8\%)\end{tabular} & \cellcolor[HTML]{FFF5F0} \begin{tabular}{@{}c@{}}+0.16 BLEU$^{\ddagger}$ \\ (-0.3\%)\end{tabular} & \cellcolor[HTML]{F1FAEE} \begin{tabular}{@{}c@{}}+0.21 BLEU$^{\ddagger}$ \\ (4.0\%)\end{tabular} & \cellcolor[HTML]{ECF8E8} \begin{tabular}{@{}c@{}}+0.16 BLEU$^{\ddagger}$ \\ (8.0\%)\end{tabular} & \cellcolor[HTML]{C3E7BC} \begin{tabular}{@{}c@{}}+0.11 BLEU$^{\ddagger}$ \\ (26.6\%)\end{tabular} \\
KDE (en$\rightarrow$es) & \cellcolor[HTML]{C6E8BF} \begin{tabular}{@{}c@{}}+0.57 BLEU$^{\ddagger}$ \\ (25.4\%)\end{tabular} & \cellcolor[HTML]{DDF2D8} \begin{tabular}{@{}c@{}}+0.23 BLEU$^{\ddagger}$ \\ (15.7\%)\end{tabular} & \cellcolor[HTML]{E3F4DE} \begin{tabular}{@{}c@{}}+0.24 BLEU$^{\ddagger}$ \\ (13.6\%)\end{tabular} & \cellcolor[HTML]{FFF2EC} \begin{tabular}{@{}c@{}}-0.01 BLEU \\ (-1.7\%)\end{tabular} & \cellcolor[HTML]{DDF2D8} \begin{tabular}{@{}c@{}}+0.12 BLEU$^{\ddagger}$ \\ (15.9\%)\end{tabular} & \cellcolor[HTML]{DAF0D4} \begin{tabular}{@{}c@{}}+0.16 BLEU$^{\ddagger}$ \\ (17.4\%)\end{tabular} & \cellcolor[HTML]{E5F5E1} \begin{tabular}{@{}c@{}}+0.04 BLEU$^{\ddagger}$ \\ (12.2\%)\end{tabular} & \cellcolor[HTML]{F0F9ED} \begin{tabular}{@{}c@{}}+0.10 BLEU$^{\ddagger}$ \\ (4.9\%)\end{tabular} \\
KDE (es$\rightarrow$en) & \cellcolor[HTML]{CFECC9} \begin{tabular}{@{}c@{}}+0.58 BLEU$^{\ddagger}$ \\ (21.8\%)\end{tabular} & \cellcolor[HTML]{E0F3DB} \begin{tabular}{@{}c@{}}+0.15 BLEU$^{\ddagger}$ \\ (14.7\%)\end{tabular} & \cellcolor[HTML]{BCE4B5} \begin{tabular}{@{}c@{}}+0.42 BLEU$^{\ddagger}$ \\ (28.7\%)\end{tabular} & \cellcolor[HTML]{FFF5F0} \begin{tabular}{@{}c@{}}+0.04 BLEU$^{\ddagger}$ \\ (-0.3\%)\end{tabular} & \cellcolor[HTML]{E1F3DC} \begin{tabular}{@{}c@{}}+0.19 BLEU$^{\ddagger}$ \\ (14.4\%)\end{tabular} & \cellcolor[HTML]{C3E7BC} \begin{tabular}{@{}c@{}}+0.28 BLEU$^{\ddagger}$ \\ (26.6\%)\end{tabular} & \cellcolor[HTML]{E2F4DD} \begin{tabular}{@{}c@{}}+0.09 BLEU$^{\ddagger}$ \\ (13.9\%)\end{tabular} & \cellcolor[HTML]{E1F3DC} \begin{tabular}{@{}c@{}}+0.32 BLEU$^{\ddagger}$ \\ (14.2\%)\end{tabular} \\
KDE (en$\rightarrow$pt) & \cellcolor[HTML]{D4EECE} \begin{tabular}{@{}c@{}}+0.52 BLEU$^{\ddagger}$ \\ (19.8\%)\end{tabular} & \cellcolor[HTML]{C9EAC2} \begin{tabular}{@{}c@{}}+0.43 BLEU$^{\ddagger}$ \\ (24.3\%)\end{tabular} & \cellcolor[HTML]{EBF7E7} \begin{tabular}{@{}c@{}}+0.27 BLEU$^{\ddagger}$ \\ (8.9\%)\end{tabular} & \cellcolor[HTML]{FFEEE6} \begin{tabular}{@{}c@{}}+0.10 BLEU$^{\ddagger}$ \\ (-4.5\%)\end{tabular} & \cellcolor[HTML]{F5FBF2} \begin{tabular}{@{}c@{}}+0.06 BLEU$^{\ddagger}$ \\ (1.7\%)\end{tabular} & \cellcolor[HTML]{F2FAF0} \begin{tabular}{@{}c@{}}+0.23 BLEU$^{\ddagger}$ \\ (3.4\%)\end{tabular} & \cellcolor[HTML]{EEF8EA} \begin{tabular}{@{}c@{}}+0.23 BLEU$^{\ddagger}$ \\ (6.4\%)\end{tabular} & \cellcolor[HTML]{B1E0AB} \begin{tabular}{@{}c@{}}+0.35 BLEU$^{\ddagger}$ \\ (32.2\%)\end{tabular} \\
KDE (pt$\rightarrow$en) & \cellcolor[HTML]{C1E6BA} \begin{tabular}{@{}c@{}}+0.40 BLEU$^{\ddagger}$ \\ (27.3\%)\end{tabular} & \cellcolor[HTML]{EDF8E9} \begin{tabular}{@{}c@{}}+0.05 BLEU$^{\ddagger}$ \\ (7.2\%)\end{tabular} & \cellcolor[HTML]{9BD696} \begin{tabular}{@{}c@{}}+0.47 BLEU$^{\ddagger}$ \\ (39.3\%)\end{tabular} & \cellcolor[HTML]{F2FAF0} \begin{tabular}{@{}c@{}}+0.28 BLEU$^{\ddagger}$ \\ (3.4\%)\end{tabular} & \cellcolor[HTML]{EFF9EB} \begin{tabular}{@{}c@{}}+0.43 BLEU$^{\ddagger}$ \\ (5.9\%)\end{tabular} & \cellcolor[HTML]{C0E6B9} \begin{tabular}{@{}c@{}}+0.21 BLEU$^{\ddagger}$ \\ (27.6\%)\end{tabular} & \cellcolor[HTML]{DAF0D4} \begin{tabular}{@{}c@{}}+0.19 BLEU$^{\ddagger}$ \\ (17.3\%)\end{tabular} & \cellcolor[HTML]{D3EECD} \begin{tabular}{@{}c@{}}+0.18 BLEU$^{\ddagger}$ \\ (20.0\%)\end{tabular} \\
NLU\_AgNews & \cellcolor[HTML]{087432}\color{white} \begin{tabular}{@{}c@{}}+0.35 Acc$^{\ddagger}$ \\ (84.7\%)\end{tabular} & \cellcolor[HTML]{55B567} \begin{tabular}{@{}c@{}}+0.00 Acc \\ (57.7\%)\end{tabular} & \cellcolor[HTML]{78C679} \begin{tabular}{@{}c@{}}+0.00 Acc \\ (48.9\%)\end{tabular} & \cellcolor[HTML]{1D8640}\color{white} \begin{tabular}{@{}c@{}}+0.00 Acc \\ (77.0\%)\end{tabular} & \cellcolor[HTML]{359E53}\color{white} \begin{tabular}{@{}c@{}}+0.00 Acc \\ (67.6\%)\end{tabular} & \cellcolor[HTML]{86CC85} \begin{tabular}{@{}c@{}}+0.00 Acc \\ (45.0\%)\end{tabular} & \cellcolor[HTML]{FFF0E8} \begin{tabular}{@{}c@{}}+0.00 Acc \\ (-3.4\%)\end{tabular} & \cellcolor[HTML]{FFF0E8} \begin{tabular}{@{}c@{}}+0.00 Acc \\ (-3.5\%)\end{tabular} \\
NLU\_CR & \cellcolor[HTML]{05712F}\color{white} \begin{tabular}{@{}c@{}}+0.90 Acc$^{\ddagger}$ \\ (85.9\%)\end{tabular} & \cellcolor[HTML]{1F8742}\color{white} \begin{tabular}{@{}c@{}}+2.10 Acc$^{\ddagger}$ \\ (76.5\%)\end{tabular} & \cellcolor[HTML]{005A24}\color{white} \begin{tabular}{@{}c@{}}+1.25 Acc$^{\ddagger}$ \\ (93.2\%)\end{tabular} & \cellcolor[HTML]{127C39}\color{white} \begin{tabular}{@{}c@{}}+0.85 Acc$^{\ddagger}$ \\ (81.0\%)\end{tabular} & \cellcolor[HTML]{005522}\color{white} \begin{tabular}{@{}c@{}}+0.80 Acc$^{\ddagger}$ \\ (94.9\%)\end{tabular} & \cellcolor[HTML]{006027}\color{white} \begin{tabular}{@{}c@{}}+0.55 Acc$^{\ddagger}$ \\ (91.1\%)\end{tabular} & \cellcolor[HTML]{005924}\color{white} \begin{tabular}{@{}c@{}}+0.85 Acc$^{\ddagger}$ \\ (93.6\%)\end{tabular} & \cellcolor[HTML]{067230}\color{white} \begin{tabular}{@{}c@{}}+0.60 Acc$^{\ddagger}$ \\ (85.5\%)\end{tabular} \\
NLU\_MNLI & \cellcolor[HTML]{A2D99C} \begin{tabular}{@{}c@{}}+1.00 Acc$^{\ddagger}$ \\ (37.3\%)\end{tabular} & \cellcolor[HTML]{A0D99B} \begin{tabular}{@{}c@{}}+1.00 Acc$^{\ddagger}$ \\ (37.8\%)\end{tabular} & \cellcolor[HTML]{DDF2D8} \begin{tabular}{@{}c@{}}+0.00 Acc \\ (16.0\%)\end{tabular} & \cellcolor[HTML]{FFF0E8} \begin{tabular}{@{}c@{}}+0.00 Acc \\ (-3.4\%)\end{tabular} & \cellcolor[HTML]{FFF0E8} \begin{tabular}{@{}c@{}}+0.00 Acc \\ (-3.5\%)\end{tabular} & \cellcolor[HTML]{84CC83} \begin{tabular}{@{}c@{}}+0.00 Acc \\ (45.3\%)\end{tabular} & \cellcolor[HTML]{FFF0E8} \begin{tabular}{@{}c@{}}+0.00 Acc \\ (-3.5\%)\end{tabular} & \cellcolor[HTML]{FFEFE8} \begin{tabular}{@{}c@{}}+0.00 Acc \\ (-3.7\%)\end{tabular} \\
NLU\_QNLI & \cellcolor[HTML]{E9F7E5} \begin{tabular}{@{}c@{}}+1.10 Acc$^{\ddagger}$ \\ (9.9\%)\end{tabular} & \cellcolor[HTML]{FFF5F0} \begin{tabular}{@{}c@{}}+0.10 Acc \\ (-0.3\%)\end{tabular} & \cellcolor[HTML]{F4FBF2} \begin{tabular}{@{}c@{}}+0.00 Acc \\ (2.0\%)\end{tabular} & \cellcolor[HTML]{006D2C}\color{white} \begin{tabular}{@{}c@{}}+1.40 Acc$^{\ddagger}$ \\ (87.2\%)\end{tabular} & \cellcolor[HTML]{D5EFCF} \begin{tabular}{@{}c@{}}+0.05 Acc \\ (19.3\%)\end{tabular} & \cellcolor[HTML]{EBF7E7} \begin{tabular}{@{}c@{}}+0.00 Acc \\ (8.4\%)\end{tabular} & \cellcolor[HTML]{077331}\color{white} \begin{tabular}{@{}c@{}}+0.05 Acc \\ (84.8\%)\end{tabular} & \cellcolor[HTML]{42AB5D}\color{white} \begin{tabular}{@{}c@{}}+1.00 Acc$^{\ddagger}$ \\ (62.2\%)\end{tabular} \\
NLU\_SST-2 & \cellcolor[HTML]{66BD6F} \begin{tabular}{@{}c@{}}+0.10 Acc \\ (53.1\%)\end{tabular} & \cellcolor[HTML]{88CE87} \begin{tabular}{@{}c@{}}+0.10 Acc \\ (44.4\%)\end{tabular} & \cellcolor[HTML]{005F26}\color{white} \begin{tabular}{@{}c@{}}+0.60 Acc$^{\ddagger}$ \\ (91.5\%)\end{tabular} & \cellcolor[HTML]{65BD6F} \begin{tabular}{@{}c@{}}+0.30 Acc$^{\ddagger}$ \\ (53.9\%)\end{tabular} & \cellcolor[HTML]{1C8540}\color{white} \begin{tabular}{@{}c@{}}+0.10 Acc \\ (77.4\%)\end{tabular} & \cellcolor[HTML]{2F974E}\color{white} \begin{tabular}{@{}c@{}}+0.45 Acc$^{\ddagger}$ \\ (70.1\%)\end{tabular} & \cellcolor[HTML]{005522}\color{white} \begin{tabular}{@{}c@{}}+0.80 Acc$^{\ddagger}$ \\ (94.8\%)\end{tabular} & \cellcolor[HTML]{72C375} \begin{tabular}{@{}c@{}}+0.10 Acc \\ (50.8\%)\end{tabular} \\
NLU\_SST-5 & \cellcolor[HTML]{3BA458}\color{white} \begin{tabular}{@{}c@{}}+3.10 Acc$^{\ddagger}$ \\ (65.2\%)\end{tabular} & \cellcolor[HTML]{98D594} \begin{tabular}{@{}c@{}}+0.05 Acc \\ (40.1\%)\end{tabular} & \cellcolor[HTML]{70C274} \begin{tabular}{@{}c@{}}+1.90 Acc$^{\ddagger}$ \\ (50.9\%)\end{tabular} & \cellcolor[HTML]{4DB163} \begin{tabular}{@{}c@{}}+1.40 Acc$^{\ddagger}$ \\ (59.6\%)\end{tabular} & \cellcolor[HTML]{46AE60}\color{white} \begin{tabular}{@{}c@{}}+2.55 Acc$^{\ddagger}$ \\ (61.0\%)\end{tabular} & \cellcolor[HTML]{29914A}\color{white} \begin{tabular}{@{}c@{}}+2.25 Acc$^{\ddagger}$ \\ (72.5\%)\end{tabular} & \cellcolor[HTML]{17813D}\color{white} \begin{tabular}{@{}c@{}}+1.50 Acc$^{\ddagger}$ \\ (78.9\%)\end{tabular} & \cellcolor[HTML]{39A257}\color{white} \begin{tabular}{@{}c@{}}+1.80 Acc$^{\ddagger}$ \\ (65.8\%)\end{tabular} \\
NLU\_Subj & \cellcolor[HTML]{3AA357}\color{white} \begin{tabular}{@{}c@{}}+2.00 Acc$^{\ddagger}$ \\ (65.3\%)\end{tabular} & \cellcolor[HTML]{FFEFE8} \begin{tabular}{@{}c@{}}+0.00 Acc \\ (-3.8\%)\end{tabular} & \cellcolor[HTML]{FFF0E8} \begin{tabular}{@{}c@{}}+0.00 Acc \\ (-3.2\%)\end{tabular} & \cellcolor[HTML]{137D39}\color{white} \begin{tabular}{@{}c@{}}+4.55 Acc$^{\ddagger}$ \\ (80.5\%)\end{tabular} & \cellcolor[HTML]{E5F5E1} \begin{tabular}{@{}c@{}}+0.00 Acc \\ (12.1\%)\end{tabular} & \cellcolor[HTML]{FFF0E8} \begin{tabular}{@{}c@{}}+0.00 Acc \\ (-3.5\%)\end{tabular} & \cellcolor[HTML]{FFEFE8} \begin{tabular}{@{}c@{}}+0.00 Acc \\ (-3.7\%)\end{tabular} & \cellcolor[HTML]{FFEEE7} \begin{tabular}{@{}c@{}}+0.00 Acc \\ (-3.9\%)\end{tabular} \\
\hline
\end{tabular}%
   }
\caption{Performance gains (BLEU/Accuracy) and token savings (\%) evaluated using the Gating (test-set calibrated) configuration. Cell colors represent the percentage of tokens saved. $^{\ddagger}$ denotes statistical significance vs.\ full reranking ($p<0.05$, two-sided).}
\label{tab:heatmap_results_test_calibrated}
\end{table*}

\paragraph{Negative savings.} A small number of cells in Tables~\ref{tab:heatmap_results_dev_calibrated} and~\ref{tab:heatmap_results_test_calibrated} show negative token savings, meaning gating triggers reranking on a large enough share of instances that its cost exceeds Full Reranking's. This is most pronounced on JRC-Acquis: Mistral-7B-Instruct-v0.3 on JRC-Acquis (en$\rightarrow$pt) is negative under dev-set calibration ($-4.9\%$, also $-8.1\%$ under test-set calibration), and JRC-Acquis (pt$\rightarrow$en) is negative for most models under dev-set calibration (e.g., Llama-3.2-3B: $-5.4\%$, also $-5.7\%$ under test-set calibration). A practitioner should check the savings for their own model-dataset combination beforehand and skip gating for any combination that comes back negative rather than deploying it blindly. Negative NLU cells are smaller in magnitude and appear mainly under test-set calibration (e.g., Llama-3.3-70B-Instruct on NLU\_Subj: $-3.9\%$).

\end{document}